%% file: iclr2026_conference.tex
\documentclass{article} 
\usepackage{iclr2026_conference,times}

\input{math_commands.tex}

\usepackage{hyperref}
\usepackage{url}
\usepackage{graphicx}
\usepackage{cleveref}
\usepackage{tcolorbox}
\usepackage{booktabs}
\usepackage{colortbl}
\usepackage{amsmath}
\usepackage{amssymb}
\usepackage{mathtools}
\usepackage{amsthm}
\usepackage{listings} 
\usepackage{xcolor}   
\usepackage{subcaption}
\usepackage{adjustbox}
\usepackage{caption}

\usepackage{wrapfig}
\usepackage{enumitem}
\usepackage{lipsum}

\definecolor{iccvblue}{rgb}{0.21,0.49,0.74}
\definecolor{lightgreen}{rgb}{0.,0.3,0.}
\definecolor{stronggreen}{rgb}{0.,0.6,0.}
\definecolor{strongred}{rgb}{0.8,0.,0.}
\definecolor{urlpink}{rgb}{1,0.4,0.5}

\definecolor{pythonblue}{RGB}{0,0,255}       
\definecolor{pythongreen}{RGB}{0,128,0}      
\definecolor{pythonred}{RGB}{200,0,0}        
\definecolor{pythonpurple}{RGB}{128,0,128}   
\definecolor{pythongreen2}{RGB}{150,15,155}

\definecolor{analysisblue}{RGB}{31,119,180}
\definecolor{analysisorange}{RGB}{255,127,14}
\definecolor{analysispurple}{RGB}{123,104,238}
\definecolor{urlpink}{rgb}{1,0.4,0.5}

\lstdefinestyle{pythonstyle}{
    commentstyle=\color{pythongreen},    
    keywordstyle=\color{pythonblue},     
    stringstyle=\color{pythonred},       
    identifierstyle=\color{black},       
    emph={print,def,return, linear_MMD},             
    emphstyle=\color{pythongreen2},      
    basicstyle=\ttfamily\normalsize,          
    breakatwhitespace=false,
    breaklines=true,
    captionpos=b,
    keepspaces=true,
    showspaces=false,
    showstringspaces=false,
    showtabs=false,
    tabsize=4
}

\title{\underline{S}cale-\underline{w}ise \underline{D}istillation of Diffusion Models}

\author{Nikita Starodubcev  \\
Yandex Research 
\And
Ilya Drobyshevskiy  \\ 
HSE University \& Yandex Research 
\And
Denis Kuznedelev \\
Yandex Research 
\And
Artem Babenko \\
Yandex Research 
\And
Dmitry Baranchuk \\
Yandex Research 
}

\iclrfinalcopy 

\newcommand{\ourmethod}{\textsc{SwD}\ }

\begin{document}

\maketitle

\input{sec/0_abstract_arxiv}    
\input{sec/1_intro_arxiv}

\input{sec/2_related_arxiv}
\input{sec/3_analysis_arxiv}

\input{sec/4_method_arxiv}
\input{sec/5_exps_arxiv}

\input{sec/6_conclusion_arxiv}

\bibliographystyle{iclr2026_conference}
\bibliography{iclr2026_conference}

\clearpage
\input{sec/7_supplementary}

\end{document}

%% file: math_commands.tex
\usepackage{amsmath,amsfonts,bm}

\def\eqref#1{equation~\ref{#1}}

\def\1{\bm{1}}

\DeclareMathAlphabet{\mathsfit}{\encodingdefault}{\sfdefault}{m}{sl}
\SetMathAlphabet{\mathsfit}{bold}{\encodingdefault}{\sfdefault}{bx}{n}



%% file: sec/0_abstract_arxiv.tex
\begin{abstract}

Recent diffusion distillation methods have achieved remarkable progress, enabling high-quality ${\sim}4$-step sampling for large-scale text-conditional image and video diffusion models. 
However, further reducing the number of sampling steps becomes more and more challenging, suggesting that efficiency gains may be better mined along other model axes. 
Motivated by this perspective, we introduce SwD, a scale-wise diffusion distillation framework that equips few-step models with progressive generation, avoiding redundant computations at intermediate diffusion timesteps. 
Beyond efficiency, SwD enriches the family of distribution matching distillation approaches by introducing a simple patch-level distillation objective based on Maximum Mean Discrepancy (MMD). 
This objective significantly improves the convergence of existing distillation methods and performs surprisingly well in isolation, offering a competitive baseline for diffusion distillation.
Applied to state-of-the-art text-to-image/video diffusion models, SwD approaches the sampling speed of two full-resolution steps and largely outperforms alternatives under the same compute budget, as evidenced by automatic metrics and human preference studies.
Project page: {\small{\url{https://yandex-research.github.io/swd}}}.

\end{abstract}






%% file: sec/1_intro_arxiv.tex
\section{Introduction}
\label{sec:intro}

Designing large-scale diffusion models (DMs) for high-resolution image and video generation is inherently challenging due to the slow sequential sampling requiring $20{-}50$ steps.
Although state-of-the-art DMs typically operate in ${\sim}8{\times}$ lower-resolution VAE latent spaces~\citep{rombach2021highresolution}, generation speed remains a significant bottleneck, especially for recent large-scale models with ${>}8$ billion parameters~\citep{sauer2024fast, flux, hidreami1technicalreport, wu2025qwenimagetechnicalreport}.
The challenge becomes several times more pronounced for video DMs~\citep{polyak2024moviegencastmedia, wan2025, kong2024hunyuanvideo}, as the latent resolution also scales across the temporal dimension.
 
Previous works have made substantial efforts in DM acceleration from different perspectives~\citep{lu2022dpm, song2020denoising, wimbauer2024cache, li2023q, jin2025pyramidal}.
One of the most successful directions is distilling DMs into few-step generators~\citep{song2023consistency, sauer2023adversarial, yin2024onestep, kim2024pagodaprogressivegrowingonestep}, where recent methods~\citep{yin2024improved, yin2025causvid} achieve state-of-the-art performance of large-scale text-conditional DMs for ${\sim}4$ steps.
Notably, these approaches generally focus on reducing the number of sampling steps while freezing other promising degrees of freedom, such as model architectures or input resolution.


Recently, \citet{rissanen2023generative, dieleman2024spectral} drew parallels between the coarse-to-fine nature of image diffusion and implicit spectral autoregression.
Specifically, the reverse diffusion process was shown to progressively predict higher spatial frequencies conditioned on previously generated lower frequencies via the input. 
This observation connects DMs to next-scale prediction models~\citep{VAR,voronov2024switti,Infinity}, which also sample higher spatial frequencies at each step, but do so explicitly via upscaling.
Therefore, these models suggest significant efficiency potential by performing most steps at lower resolutions.
However, despite this connection, state-of-the-art few-step DMs still operate at a fixed resolution throughout the diffusion process, highlighting an underexplored direction for improving their efficiency.

\textbf{Contribution.} 
Since most state-of-the-art DMs belong to the latent diffusion family~\citep{rombach2021highresolution}, we first address whether the spectral autoregression perspective also applies to latent representations.
In this work, we conduct a spectral analysis of existing VAE latent spaces and also extend it to the video domain.
Our findings confirm that both spatial and temporal latent resolutions implicitly increase over the diffusion process, similarly to the natural images.
This suggests that latent DMs can avoid redundant computations at intermediate noisy timesteps, where high frequencies are largely suppressed.

Motivated by this observation, we introduce a \textit{\textbf{S}cale-\textbf{w}ise \textbf{D}istillation} (SwD) framework, which transforms an arbitrary pretrained DM into a single few-step model that progressively increases spatial and temporal sample resolutions at each generation step.
SwD integrates seamlessly with existing distribution matching distillation approaches~\citep{sauer2023adversarial, yin2024onestep} and leverages their few-step sampling algorithms, which appear naturally aligned with progressive generation.

In addition to the scale-wise distillation framework, we present a simple yet surprisingly effective diffusion distillation objective that minimizes Maximum Mean Discrepancy (MMD)~\citep{gretton2012kernel} in the feature space of a pretrained DM. 
The proposed objective complements state-of-the-art distillation methods and achieves strong performance even in isolation, establishing a competitive baseline for DM distillation.
Importantly, it requires no additional trainable models, making it computationally efficient and easy to combine with existing distillation pipelines.

We apply \ourmethod to state-of-the-art text-to-image and video DMs and show that our models compete or even outperform their teachers being more than $10{\times}$ faster. 
Compared to full-resolution few-step models, \ourmethod significantly surpasses them under a similar computational budget.
For the same number of sampling steps, \ourmethod provides ${\sim}2{\times}$ speedup in text-to-image generation and ${\sim}3{\times}$ in text-to-video generation, without compromising quality.

%% file: sec/2_related_arxiv.tex
\section{Related Work}
\label{sec:background}



\textbf{Diffusion distillation into few-step models.}
Diffusion distillation methods aim to reduce generation steps to $1{-}4$ while maintaining teacher model performance. 
These methods can be largely grouped into two categories: \textit{teacher-following} methods~\citep{meng2023distillation, song2023consistency, luo2023latent, huang2023knowledge, song2024improved, kim2025autoregressive} and \textit{distribution matching}~\citep{yin2024onestep, yin2024improved,sauer2023adversarial, sauer2024fast, luo2023diffinstruct, zhou2024score, zhou2024long}.

Teacher-following methods approximate the teacher's noise-to-data mapping by integrating the diffusion ODE in fewer steps than numerical solvers~\citep{song2020denoising, lu2022dpm}.
Distribution matching methods relax the teacher-following constraint, focusing instead on aligning student and teacher distributions without requiring exact noise-to-data mapping.
State-of-the-art approaches, such as DMD2~\citep{yin2024improved} and ADD~\citep{sauer2023adversarial, sauer2024fast}, demonstrate strong generative performance in ${\sim}4$ steps. 
However, they still exhibit noticeable quality degradation at $1{-}2$ step generation, leaving room for further improvement.
Recently, DMD has been successfully adopted for video diffusion models~\citep{yin2025causvid, huang2025selfforcing}.



\textbf{Progressive generation with DMs.} 
The idea of progressively increasing resolution during diffusion generation was initially exploited in hierarchical or cascaded DMs~\citep{ho2021cascaded, saharia2022photorealistic, ramesh2022hierarchicaltextconditionalimagegeneration, kastryulin2024yaart,gu2023matryoshka}, which are strong competitors to latent DMs~\citep{rombach2021highresolution} for high-resolution generation. 
Cascaded DMs consist of multiple DMs operating at different resolutions, where each model performs a diffusion sampling from scratch, conditioned on the lower-resolution sample. 
To bridge progressive generation with diffusion processes, several works~\citep{gu2023fdm,teng2023relay,atzmon2024edify, jin2025pyramidal,zhang2025training, anagnostidis2025flexidit, haji-ali2025improving} have presented multi-stage pipelines, where DMs are trained to smoothly transition to higher-resolution noisy samples during diffusion sampling.
\ourmethod follows up this line of research by proposing a framework that readily integrates into existing diffusion distillation procedures and adapts arbitrary pretrained DMs into progressive few-step models.

\textbf{Maximum Mean Discrepancy in generative modeling.}
Maximum Mean Discrepancy (MMD) is a metric between two distributions $P$ and $Q$, widely explored in early GAN works~\citep{binkowski2018demystifying, wang2018improving, dziugaite2015training, bellemare2017cramer, sutherland2016generative}.

Given a positive-definite \textit{kernel function} $k(\mathbf{x}, \mathbf{y})$, the MMD can be defined as
\begin{equation}
\begin{gathered}
\mathrm{MMD}^2(P, Q) 
= \mathbb{E}_{\mathbf{x}, \mathbf{x}' \sim P}\!\left[ k(\mathbf{x}, \mathbf{x}') \right] 
+ \mathbb{E}_{\mathbf{y}, \mathbf{y}' \sim Q}\!\left[ k(\mathbf{y}, \mathbf{y}') \right] 
- 2 \, \mathbb{E}_{\mathbf{x} \sim P, \, \mathbf{y} \sim Q}\!\left[ k(\mathbf{x}, \mathbf{y}) \right],
\end{gathered}
\end{equation}
where $\mathbf{x}$ and $\mathbf{y}$ denote samples from the generated and target distributions, respectively. 

Generative Moment Matching Networks (GMMNs)~\citep{li2015generative} employ the MMD with a fixed Gaussian kernel (RBF) directly in data space. 
GANs, in contrast, typically consider learnable kernels, designed as the composition of a discriminator with a fixed kernel.

In diffusion modeling, the MMD has been explored for DM training~\citep{bortoli2025distributional} or finetuning~\citep{aiello2023fast}.
DMMD~\citep{galashov2025deep} employs noise-adapted discriminators for MMD gradient flows~\citep{arbel2019mmdflow}.
Recently, IMM~\citep{zhou2025inductive} leveraged the MMD for consistency distillation~\citep{song2023consistency}, computing the MMD with a fixed kernel between raw generator predictions at different timesteps.
In our work, we adopt the MMD between student and teacher distributions in the feature space of a pretrained DM, yielding a powerful and effective distribution matching objective. 

%% file: sec/3_analysis_arxiv.tex
\begin{figure}[t!]
\hspace{-2mm}
\includegraphics[width=1.02\linewidth]{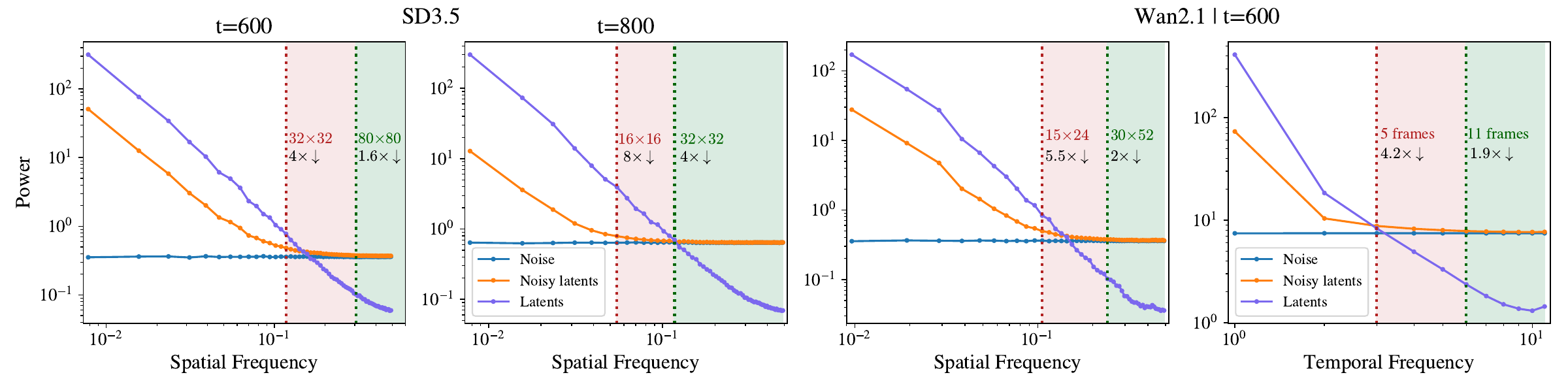}
\vspace{-6mm}
\caption{
    \textbf{Spectral analysis} of SD3.5 VAE latents ($128{\times}128$) (\textit{Left}) and Wan2.1 ($21{\times}60{\times}104$) for spatial and temporal dimensions (\textit{Right}).
    Vertical lines mark the frequency boundaries for which the frequency components to the right are not present in lower resolution latents.
    Noise masks high frequencies, suggesting that latent DMs can operate at lower latent resolutions for high noise levels.
    \textcolor{stronggreen}{Green} area indicates the allowed latent resolution at corresponding timestep, while \textcolor{strongred}{Red} area shows that further resolution reduction would lead to noticeable information loss.
}
\label{fig:anal_freqs}
\end{figure}

\section{Latent space spectral analysis}
\label{sec:analysis}


\citet{rissanen2023generative} and \citet{dieleman2024spectral} showed that, in pixel space, diffusion models approximate spectral autoregression for natural images. 
Since state-of-the-art text-conditional diffusion models operate on VAE latent representations~\citep{rombach2021highresolution}, we first investigate this spectral perspective for various latent spaces and also extend it to the video domain. 

Following \cite{dieleman2024spectral}, we evaluate \textit{radially averaged power spectral density} (RAPSD), i.e., the averaged spectra power across different spatial frequency components, and its one-dimensional analogue for temporal frequencies in video.

We examine the latent spaces of image and video diffusion models, specifically Stable Diffusion 3.5 (SD3.5)~\citep{esser2024scaling} and Wan2.1~\citep{wan2025}.
The SD3.5 VAE maps $3{\times}1024{\times}1024$ images into $16{\times}128{\times}128$ latents, while the Wan2.1 VAE encodes $81{\times}3{\times}480{\times}832$ video inputs into $21{\times}16{\times}60{\times}104$ latents. 
Both models use a flow-matching process~\citep{lipman2023flow}. 



\Cref{fig:anal_freqs} shows the RAPSD of Gaussian noise (\textcolor{analysisblue}{blue}), clean latents (\textcolor{analysispurple}{purple}) and noisy latents (\textcolor{analysisorange}{orange}) at different timesteps. 
\Cref{fig:anal_freqs} (Left) provides the results for SD3.5 VAE latents.
\Cref{fig:anal_freqs} (Right) shows RAPSD across both spatial and temporal frequencies of Wan2.1 latents.
Vertical lines indicate frequency boundaries: the components to the right correspond to high frequencies absent at lower resolutions, while those to the left align with the full latent resolution ($128{\times}128$).

Additional results, including more timesteps and SDXL~\citep{podell2024sdxl} latents under a variance-preserving diffusion process~\citep{ho2020denoising, song2020score}, are provided in Appendix~\ref{app:extended_anal}.

\textbf{Observations.}
First, we note that the latent frequency spectrum approximately follows a power law, similar to natural images~\citep{Schaaf1996ModellingTP}.
In contrast, however, highest frequency components in latent space exhibit slightly greater magnitude. 
We attribute this to the VAE regularization terms, which may cause ``clean'' latents to appear slightly noisy.

We also observe that the noising process progressively filters out high frequencies, thereby determining the safe downsampling range without noticeable information loss.
~\Cref{fig:anal_freqs} (Left) shows that at $t{=}800$, noise masks high frequency components emerging at resolutions above $32{\times}32$.
This allows for $4{\times}$ downsampling of $128{\times}128$ latents (\textcolor{stronggreen}{green} area).
On the other hand, $8{\times}$ downsampling would corrupt the data signal (\textcolor{strongred}{red} area), as the noise does not fully suppress those frequencies.

A similar effect is observed along the temporal dimension in Wan2.1 latents, see~\Cref{fig:anal_freqs} (Right).
At $t{=}600$, the effective signal can be represented with ${\sim}11$ latent frames instead of the original $21$.

\textbf{Practical implication.} 
Based on this analysis, we suppose that latent diffusion models may operate at lower resolution at high noise levels without losing the data signal.
In other words, modeling high frequencies at timestep $t$ is unnecessary if those frequencies are already masked at that noise level.
Note that this holds true for both spatial and temporal axes for video DMs. 
We summarize this conclusion as follows:
\begin{tcolorbox}[colback=green!5!white,colframe=black!75!black]
Latent diffusion allows lower-resolution modeling at high noise levels across spatial and temporal dimensions.
\end{tcolorbox}

%% file: sec/4_method_arxiv.tex
\section{Method}
\label{sect:method}

This section introduces a \textit{scale-wise distillation}, SwD, framework for diffusion models. 
We begin by describing the \ourmethod pipeline, highlighting its key features and challenges.
Then, we present our distillation objective based on Maximum Mean Discrepancy (MMD).

\subsection{Scale-wise Distillation of DMs}
\label{sect:method_scalewise}


The core design principle of \ourmethod is to unify multi-scale generation within a \textit{single few-step model} and \textit{single diffusion process}, in contrast to cascaded approaches.
To this end, we define a few-step \textit{timestep schedule}, $[t_{1}, \dots, t_{N}]$, and pair each diffusion timestep $t_{i}$ with a latent resolution $s_{i}$ from a non-decreasing \textit{scale schedule}, $[s_{1}, \dots, s_{N}]$. 
Therefore, starting the generation with Gaussian noise at the lowest scale, $s_1$, the resolution of intermediate noisy latents $\mathbf{x}_{t_i}$ is progressively increased over sampling steps.

\textbf{Upsampling strategy.} 
We first address \textit{how to upsample intermediate $\mathbf{x}_{t_i}$ to obtain faithful higher-resolution noisy samples?}
A naive approach is to directly upsample $\mathbf{x}_{t_i}$.
However, this distorts the noise variance and introduces local noise correlations. 
Consequently, recent progressive DMs~\citep{jin2025pyramidal, atzmon2024edify} propose dedicated techniques to handle jump points and preserve sampling continuity.

In contrast, few-step models enable \textit{stochastic multistep sampling}~\citep{song2023consistency} by predicting a clean sample $\hat{\mathbf{x}}_{0}$ and then renoising it to the next noise level. 
This approach naturally suggests upsampling $\hat{\mathbf{x}}_{0}$ prior to renoising, thereby preserving the correct noise statistics at higher resolution.



\input{tables/upscaling_strategies_new}

To validate this intuition, we generate images with Stable Diffusion 3.5~\citep{esser2024scaling} from intermediate noisy latents, $\mathbf{{x}}_{t}$, obtained with different upsampling strategies.
Specifically, given a full-resolution ($128{\times}128$) real image latent, $\mathbf{{x}}_{0}$, and its downscaled version ($64{\times}64$), $\mathbf{{x}}^{\text{down}}_{0}$, we consider: 
(\textbf{A}) a reference setting where noise is added to the original $\mathbf{{x}}_{0}$; 
(\textbf{B}) upsampling $\mathbf{x}^{\text{down}}_{0}$ followed by noise injection;
(\textbf{C}) injecting noise into $\mathbf{x}^{\text{down}}_{0}$ followed by upsampling.


As shown in~\Cref{tab:method_upsampling}, the naive strategy (\textbf{C}) produces largely out-of-distribution (OOD) noisy latents. 
In contrast, strategy (\textbf{B}) shows comparable results to the original noisy latents at high noise levels, likely because heavy noise masks interpolation artifacts.
The results for other upscale ratios are presented in~\Cref{app:extended_upscaling_strategies}.



\begin{figure}[t!]
\begin{minipage}{0.45\textwidth}
    \centering
    \includegraphics[width=\linewidth]{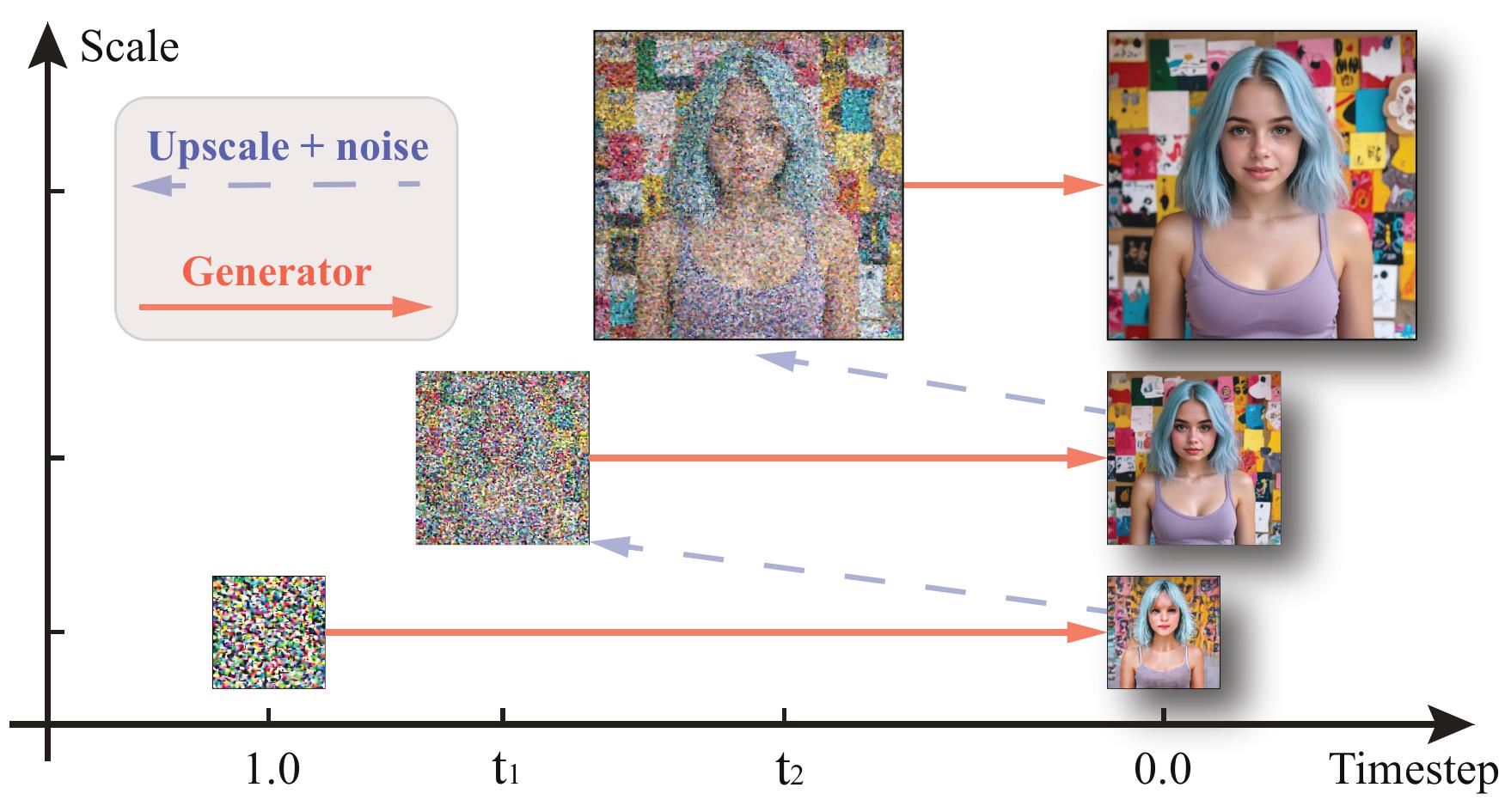}
    \vspace{-5mm}
    \caption{
        \textbf{\ourmethod sampling.}
        Few-step model starts sampling from noise at the low resolution $s_{1}$ and gradually increases it over generation steps.
        At each step, the previous denoised prediction at the scale $s_{i-1}$ is upsampled and noised according to the \textit{timestep schedule}, $t_{i}$. 
        Then, the generator predicts a clean image at the current resolution $s_{i}$.
    }
    \label{fig:method_inference}
\end{minipage}
\hfill
\begin{minipage}{0.52\textwidth}
    \vspace{-2.5mm}
    \centering
    \includegraphics[width=\linewidth]{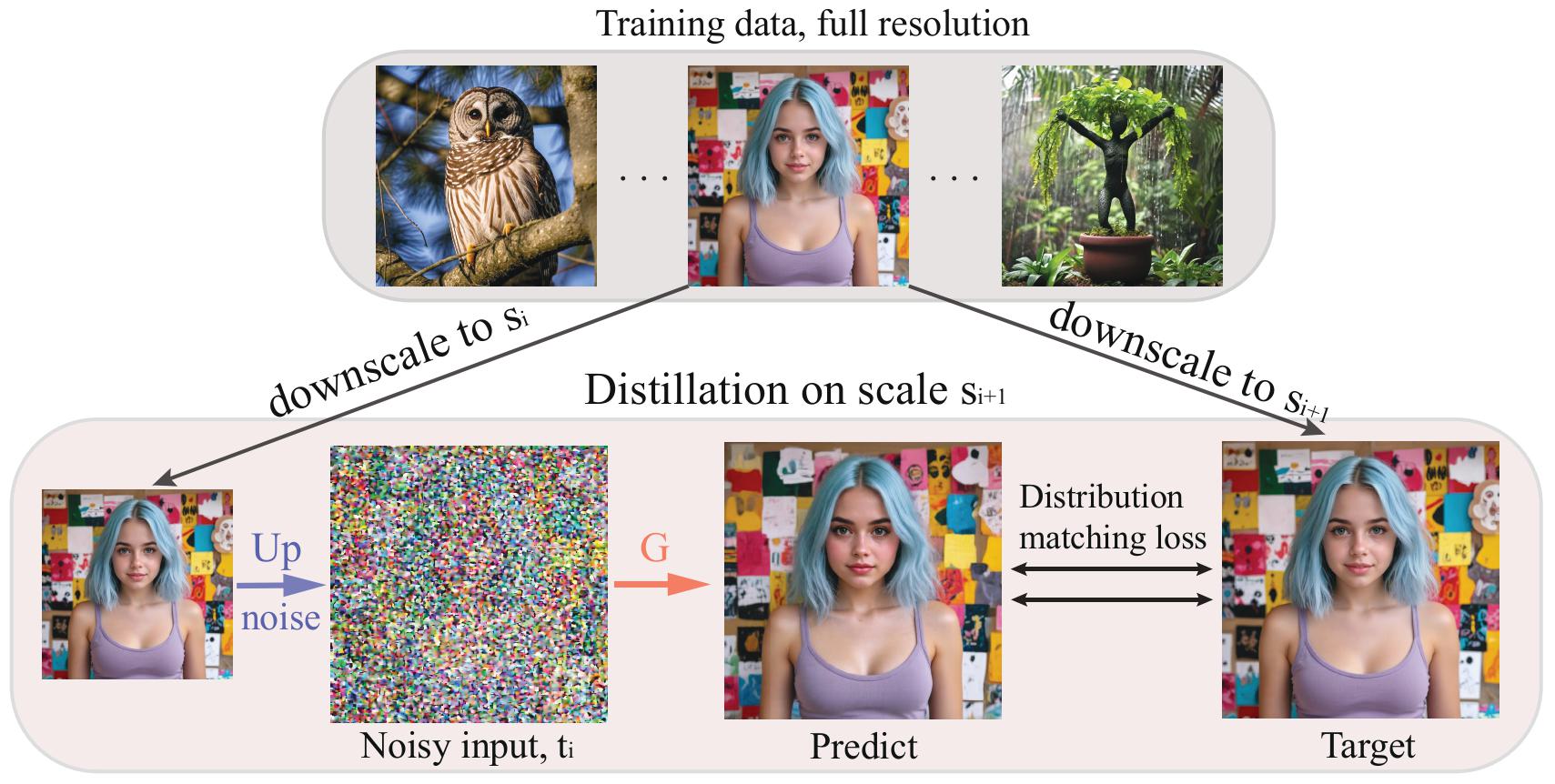}
    \vspace{-5mm}
    \caption{
        \textbf{\ourmethod training step.} 
        \textbf{i)} Sample a pair of adjacent resolutions [$s_{i}$, $s_{i+1}$] from \textit{scale schedule}.
        \textbf{ii)} Downscale the training images to $s_{i}$ and $s_{i+1}$. \textbf{iii)} The lower scale versions are upsampled and noised to a timestep $t_{i}$ with the forward process.
        \textbf{iv)} Given the noised images, the model $G$ predicts clean data at target scale $s_{i+1}$.
        \textbf{v)} Distribution matching loss is calculated between predicted and target images.
    }
    \label{fig:method_training}
\end{minipage}
\vspace{-2mm}
\end{figure}


Overall, we propose to handle scale-transition points by upsampling $\hat{\mathbf{x}}_0$ predictions, followed by renoising. 
In practice, we use \textit{bicubic interpolation} for spatial dimensions and \textit{adjacent frame blending} for the temporal one. 

\textbf{Sampling.}
\ourmethod performs \textit{progressive stochastic multistep sampling} under the discussed scale-transition approach.
Specifically, given the intermediate noisy sample $\mathbf{\hat{x}}_{t_{i-1}}$ at resolution $s_{i-1}$, the model predicts a denoised sample $\hat{\mathbf{x}}^{i-1}_{0}$.
To proceed to the next timestep $t_{i}$, $\hat{\mathbf{x}}^{i-1}_{0}$ is upsampled to $s_{i}$ and renoised with the forward diffusion process, resulting in a higher-resolution sample $\mathbf{\hat{x}}_{t_i}$.
Then, the model predicts the next $\hat{\mathbf{x}}^i_{0}$.
~\Cref{fig:method_inference} illustrates this sampling process.

Although this procedure can be applied directly to pretrained few-step models, we observe in practice that renoising is insufficient to completely remove upsampling artifacts, unless very high noise levels are used.
Therefore, we aim to train a few-step generator that also serves as a robust upscaler.


 

\textbf{Training.}
We train a single model across multiple resolutions, iterating over pairs of adjacent scales [$s_{i}$, $s_{i+1}$] from the \textit{scale schedule}. At each training step, we sample a batch of full-resolution images or videos, downscale them to the source and target resolutions in pixel space, according to the $s_{i}$ and $s_{i+1}$ scales, and then encode them into the VAE latent space. 
Notably, we find that downscaling in pixel space before the VAE encoding largely outperforms latent downscaling in our experiments. 

Next, we upsample the lower resolution latents from $s_{i}$ to $s_{i+1}$ and apply the forward diffusion process according to the \textit{timestep schedule}, $t_{i}$. 
The noised latents are then fed into the scale-wise generator, which predicts
${{\mathbf{\hat{x}}}}_0$ at the target scale $s_{i+1}$.

Finally, we calculate a distillation loss between the predicted and target latents at $s_{i+1}$.
In our work, we use distribution matching, motivated by ADD~\citep{sauer2023adversarial,sauer2024fast} and DMD~\citep{yin2024onestep, yin2024improved}, achieving state-of-the-art performance in diffusion distillation.

The schematic illustration of this training procedure is provided in Figure~\ref{fig:method_training}.
Further implementation details and discussions are in Appendix~\ref{app:implement_details}.




\textbf{Discussion on the timestep and scale schedules.}
Following~\Cref{sec:analysis}, the emergence of higher-frequency components at lower noise levels can provide useful initial assumptions for designing the schedules. 
However, since the analysis provides only averaged results and does not account for upsampling artifacts, the schedules ultimately remain hyperparameters.

In practice, we adopt default few-step timestep schedules and slightly shift them toward noisier timesteps, aligning with the intuition that noise injection helps mitigate upsampling artifacts.
The scale schedules are progressively increasing, starting with $2{-}4{\times}$ lower resolution to achieve the desired inference speedups.
We also find that the method is not highly sensitive to specific schedule choices, allowing them to be selected with minimal tuning and shared across models with similar diffusion processes.

\subsection{Diffusion Distillation with Maximum Mean Discrepancy}
\label{sect:method_mmd}
In addition to the proposed scale-wise distillation framework, we extend the family of distribution matching distillation methods with a 
patch-level MMD loss, calculated on the intermediate spatial features of the pretrained DMs.
Below, we discuss the loss computation for transformer-based DMs~\citep{Peebles2022DiT}, a dominant architecture in state-of-the-art DMs, and note that its formulation readily extends to convolutional backbones. 

First, we leverage the ability of DMs to operate at different noise levels, enabling the extraction of structured and finer-grained signals at high and low noise levels, respectively.
Accordingly, before feature extraction, we 
noise both generated and target samples within a predefined timestep interval.
In practice, we find that a low-to-mid noise interval yields slightly better performance.

Then, we extract feature maps $\mathbf{F}{\in}\mathbb{R}^{N{\times}L{\times}C}$ from the middle transformer block of the teacher DM for generated and target samples and denote them as $\mathbf{F}^{\text{fake}}$ and $\mathbf{F}^{\text{real}}$, respectively. 
$N$ is a batch size, $L$ is a number of spatial tokens, and $C$ is a hidden dimension of the transformer.
We then compute the MMD between the distributions of spatial tokens, which correspond to patch representations in vision transformers~\citep{dosovitskiy2020image}.

For the MMD computation, we consider two kernels: linear ($k(\mathbf{x}, \mathbf{y}) = \mathbf{x}^T\mathbf{y}$) and the radial basis function (RBF)~\citep{chang2010training}. 
The former aligns feature distribution means, while the latter also matches all higher-order moments.
In our experiments, both kernels perform similarly, so we simplify $\mathcal{L}_{\text{MMD}}$ using the linear kernel, i.e., calculate MSE between spatial token means \textit{per image}:
\begin{equation}
    \begin{gathered}
        \mathcal{L}_{\text{MMD}} = \sum^N_{n=1} \left\lVert \frac{1}{L}\sum^{L}_{l=1} \mathbf{F}^{\text{real}}_{n,l,\cdot} - \frac{1}{L}\sum^{L}_{l=1} \mathbf{F}^{\text{fake}}_{n,l,\cdot} \right\rVert ^{2}.
    \end{gathered}
    \label{eq:mmd_loss}
\end{equation}
Note that the feature means computed across the entire batch rather than per image tend to mitigate condition-specific information, resulting in lower text relevance in our experiments.

\textbf{Discussion.} 
$\mathcal{L}_{\text{MMD}}$ with a linear kernel can be considered as a diffusion distillation adaptation of the feature matching loss, proposed for improved GAN training~\citep{salimans2016improved}.
To our knowledge, such losses have not been explored in the context of diffusion distillation.
The notable differences are: i) $\mathcal{L}_{\text{MMD}}$ leverages a pretrained DM instead of a learnable discriminator; ii) it uses the feedback from different noise levels; iii) the feature means are computed per image rather than across the entire batch.





\textbf{Overall objective.}
We incorporate $\mathcal{L}_{\text{MMD}}$ as an additional loss to $\mathcal{L}_{\text{DMD}}$ and $\mathcal{L}_{\text{GAN}}$ in our scale-wise framework:  $\mathcal{L}_{\text{SwD}} = \mathcal{L}_{\text{MMD}} + \alpha\cdot\mathcal{L}_{\text{DMD}} + \beta\cdot\mathcal{L}_{\text{GAN}}$.
Interestingly, despite its simplicity, $\mathcal{L}_{\text{MMD}}$ is later shown to be a highly competitive standalone distillation objective.

%% file: tables/upscaling_strategies_new.tex
\begin{wraptable}{l}{0.5\textwidth}
\vspace{-3mm}
\centering
\resizebox{0.5\textwidth}{!}{%
\begin{tabular}{@{} l *{5}{c} @{}}
\toprule
Configuration  & $t=400$  & $t=600$ & $t=800$ \\
\midrule
    \rowcolor[HTML]{eeeeee} \textbf{A} \ $\mathbf{{x}}_{0}\xrightarrow[]
    {\text{noise}}\mathbf{{x}}_{t}$ & $9.2$ & $10.3$ & $13.7$ \\
    \textbf{B} \ $\mathbf{{x}}^{\text{down}}_{0}\xrightarrow[]{\text{upscale}}\mathbf{{x}}_{0}\xrightarrow[]{\text{noise}}\mathbf{{x}}_{t}$ & 
    $30.9$ & $18.8$ & $14.7$ \\
    \textbf{C} \ $\mathbf{{x}}^{\text{down}}_{0}\xrightarrow[]{\text{noise}}\mathbf{{x}}^{\text{down}}_{t}\xrightarrow[]{\text{upscale}}\mathbf{{x}}_{t}$  & $122$ & $223$ & $327$ \\
\bottomrule
    
\end{tabular}}
\vspace{-2mm}
\caption{
    Comparison of noisy latent upsampling strategies (\textbf{B}, \textbf{C}) for $64{\rightarrow}128$ in generation quality (FID-5K). 
    Upsampling $\mathbf{x}^{\text{down}}_0$ before noise injection (\textbf{B}) aligns better with original full-resolution noisy latents (\textbf{A}). 
}
\vspace{-3mm}
\label{tab:method_upsampling}
\end{wraptable}

%% file: sec/5_exps_arxiv.tex
\section{Experiments}
\label{sec:exps}

\textbf{Models.} 
We validate our approach in text-to-image generation by distilling SDXL~\cite{podell2024sdxl}, SD3.5 Medium, SD3.5 Large~\citep{esser2024scaling} and FLUX.1-dev~\citep{flux}. 
We also apply \ourmethod to the recent text-to-video model, Wan2.1-1.3B~\citep{wan2025}.

\textbf{Data.}
To remain in the isolated distillation setting and avoid biases from external data, we train all models exclusively on synthetic data generated by their teacher, rather than on real data. 
We note that this step does not pose a bottleneck for training, as the distillation process itself converges relatively fast (${\sim}3$K iterations) and requires significantly less data than the DM training.
The synthetic data generation settings for each model are provided in Appendix~\ref{app:implement_details}.

\textbf{Metrics.} 
For text-to-image models, we use $30$K text prompts from the COCO2014 and MJHQ sets~\citep{ms_coco, li2024playground} and evaluate the automatic metrics: FID~\citep{heusel2017gans}, HPSv3~\citep{ma2025hpsv3}, ImageReward (IR)~\citep{xu2023imagereward}, and PickScore (PS)~\citep{Kirstain2023PickaPicAO} and GenEval~\citep{geneval}.
Note that FID was shown to correlate poorly with human perception~\citep{Kirstain2023PickaPicAO} for text-to-image assessment but we report it here for completeness.

Also, we conduct a user preference study via side-by-side comparisons evaluated by professional assessors. 
We select $128$ text prompts from the PartiPrompts dataset~\citep{yu2022scaling}, following~\citep{yin2024improved, sauer2024fast}, and generate $2$ images per prompt. 
More details are in Appendix~\ref{app:human_eval}.


For T2V models, we evaluate VBench-2.0~\citep{zheng2025vbench2}, and VisionReward~\citep{xu2024visionrewardfinegrainedmultidimensionalhuman} and VideoReward~\citep{liu2025improving} on $1003$ prompts from MovieGenBench~\citep{polyak2024movie}.

\begin{figure*}[t!]
    \centering
    \includegraphics[width=0.97\linewidth]{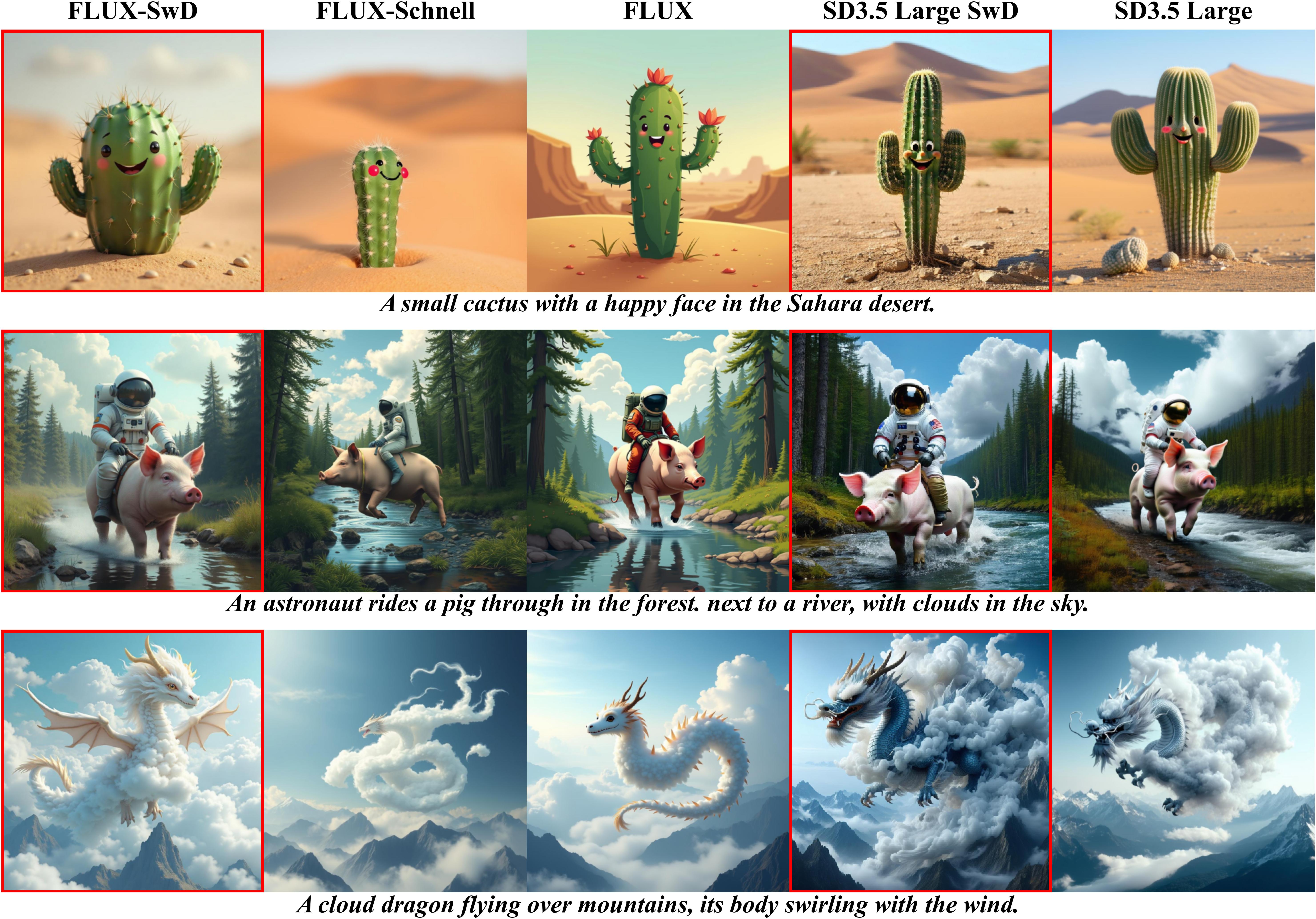}
    \vspace{-2mm}
    \caption{
        Qualitative results of FLUX-SwD and SD3.5 Large SwD.
        More examples are in~\Cref{fig:exps_cherry_app}.
    }
    \label{fig:exps_cherry}
\end{figure*}
\begin{figure*}
\vspace{-2mm}
    \centering
    \includegraphics[width=1.0\linewidth]{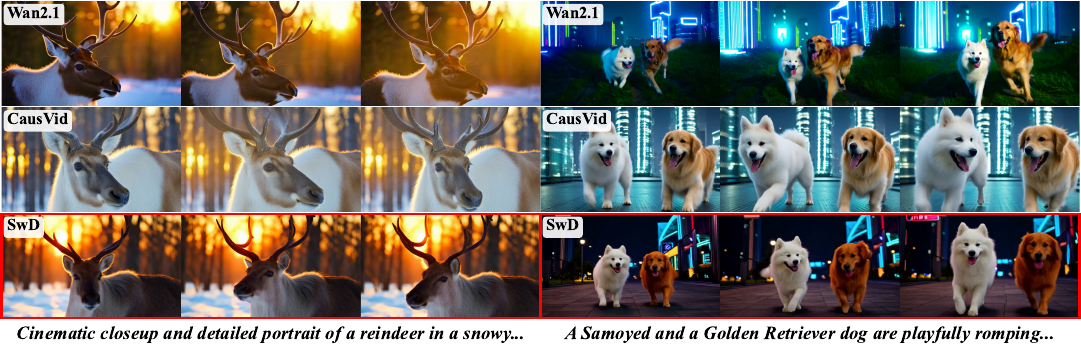}
    \vspace{-7mm}
    \caption{
        Qualitative results of Wan2.1-SwD. 
        More examples are in~\Cref{fig:app_cherry_video}.
    }
    \label{fig:exps_cherry_video}
\end{figure*}

\textbf{Setup.} 
In our main experiments, we distill the models to $4$ or $6$ steps.
For text-to-image models, the scale schedules begin at image resolutions of $256{\times}256$ or $512{\times}512$ and progress to $1024{\times}1024$.
For text-to-video, we start with $21{\times}160{\times}272$ and achieve the $81{\times}480{\times}832$ resolution. 
We chose such starting points as lower resolutions provide only marginal speed improvements. 
The exact timestep and scale schedules for each model are in Appendix~\ref{app:main_setups}.

Also, we note that SDXL and Wan2.1 are distilled using only the MMD loss, as the DMD loss performs poorly in the scale-wise setting when the base model fails to generate plausible low-resolution images. 
We discuss this limitation in more detail in Appendix~\ref{app:scale_adapted_teacher}.


\textbf{Baselines.} 
For text-to-image, we compare with the teacher models and their publicly available distilled versions: DMD2-SDXL~\citep{yin2024improved}, SDXL-Turbo~\citep{sauer2023adversarial}, Hyper-SD~\citep{ren2024hypersd}, SD3.5-Turbo~\citep{sauer2024fast}, FLUX-Schnell~\citep{flux}, FLUX-Turbo-Alpha~\citep{FLUX.1TurboAlpha}.
Also, we evaluate other fast state-of-the-art models, such as next-scale prediction models (Switti~\citep{voronov2024switti} and Infinity~\citep{Infinity}).

For the text-to-video task, we compare with the teacher model (Wan2.1-1.3B~\citep{wan2025}) and its $3$-step DMD distilled variant, CausVid~\citep{yin2025causvid}.
To ensure a fair comparison, we train CausVid using the official code and setup, but on our training dataset generated with Wan2.1-1.3B, whereas the public version uses high-quality internal data.

\subsection{Main results}
\label{sec:main_results}

\begin{figure*}
    \centering
    \hspace{-1mm}\includegraphics[width=1.01\linewidth]{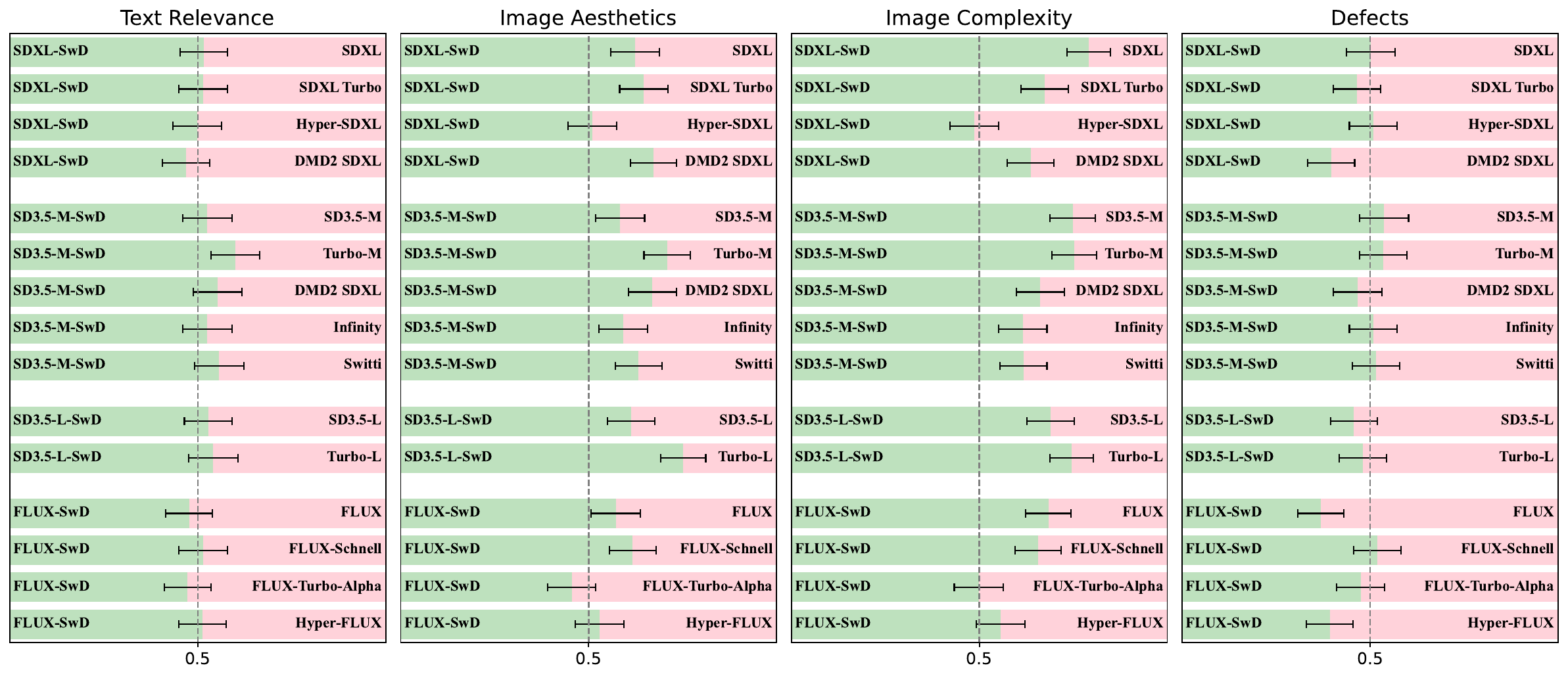}
    \vspace{-6mm}
    \caption{
        Human preference study for \ourmethod against the baseline models.
    }
    \label{fig:exps_sbs_baselines}
\end{figure*}

\textbf{Text-to-image.}
Table~\ref{tab:exps_main_results} and Figure~\ref{fig:exps_sbs_baselines} present the comparisons of \ourmethod with the baselines in terms of generation quality and speed. 
The results are organized into subsections corresponding to different base diffusion models. 

We find that \ourmethod models achieve state-of-the-art performance in terms of PS, HPSv3, IR and GenEval within their respective model families. 
In terms of latency, \ourmethod shows nearly $2{\times}$ speedup compared to the fastest counterparts.

According to the human study, \ourmethod outperforms most other models in terms of \textit{image complexity} and \textit{image aesthetics}, including the more expensive teachers and their distilled variants, while maintaining comparable levels of \textit{text relevance} and \textit{defects}.
We observe a slight degradation in defects for FLUX-SwD compared to the Hyper-FLUX and FLUX models, which are ${\sim}4{\times}$ and ${\sim}14{\times}$ slower, respectively.
SDXL-SwD also has slightly more defects than SDXL-DMD2, which we attribute to the use of the MMD loss alone, as discussed in~\Cref{sec:ablate_mmd}.
Qualitative comparisons are presented in Figure~\ref{fig:exps_cherry}, with additional results in Figures~\ref{fig:exps_cherry_app}, ~\ref{fig:app_cherry}, and~\ref{fig:app_sdxl_cherry}.

\input{tables/main_table_video}
\textbf{Text-to-video.}
The results in~\Cref{tab:exps_main_results_video} show that \ourmethod achieves better performance than the teacher model, while being $72{\times}$ faster.
Compared to CausVid, SwD provides comparable results and achieves ${\sim}2.3\times$ faster inference.
Also, we find that SwD, when applied across both temporal and spatial dimensions, does not degrade quality compared to the spatial-only variant and further improves inference speed.
Visual examples are provided in Figures~\ref{fig:exps_cherry_video} and ~\ref{fig:app_cherry_video}.

\input{tables/main_table_hpsv3}

\subsection{Scale-wise versus Full-resolution}
\label{sec:scale_vs_full}

Next, we compare \ourmethod against their full-resolution counterparts. 
The full-scale baselines use the same timestep schedules but operate at a fixed target latent resolution.

\begin{figure}
\centering
\includegraphics[width=0.47\linewidth]{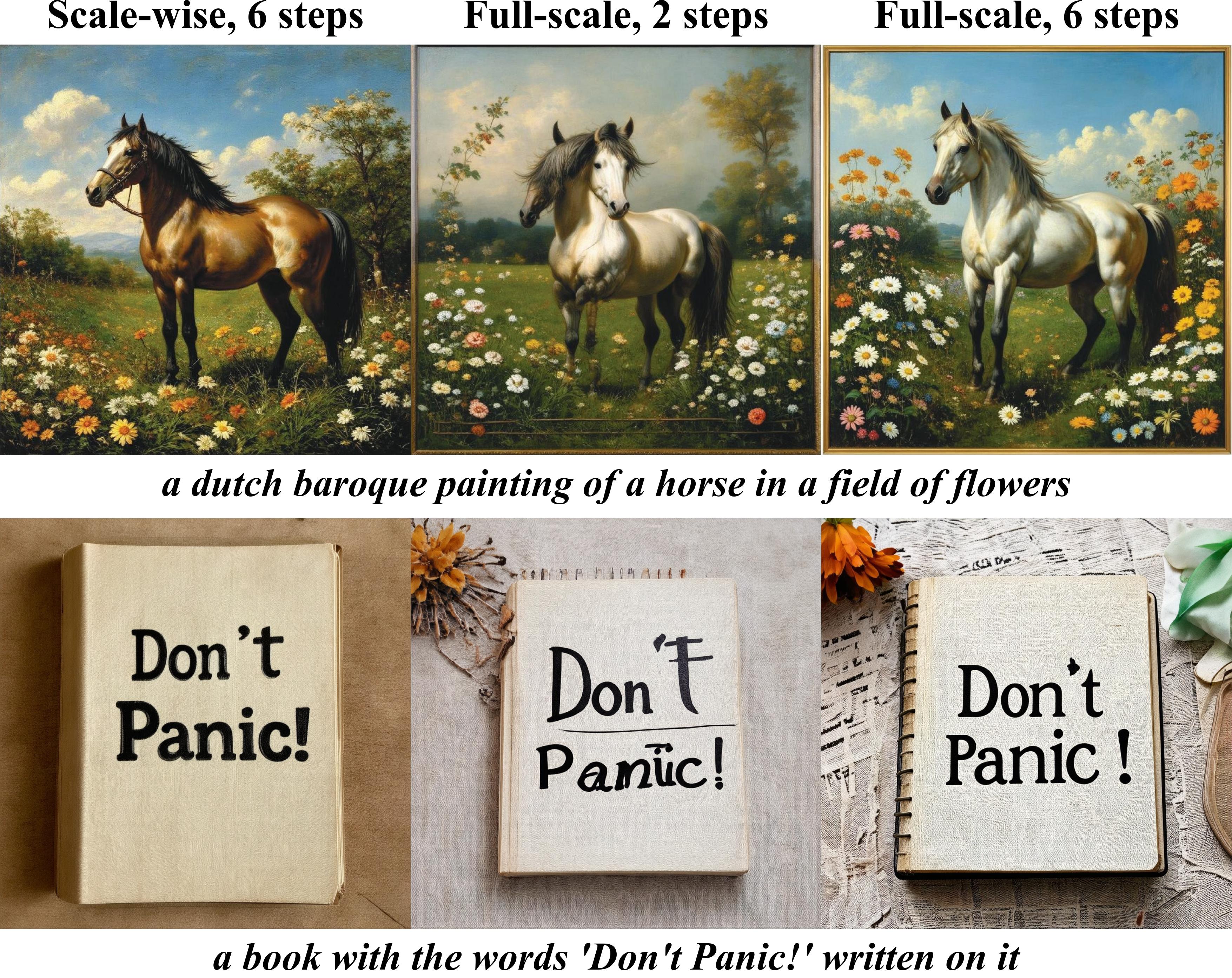}
\includegraphics[width=0.52\linewidth]{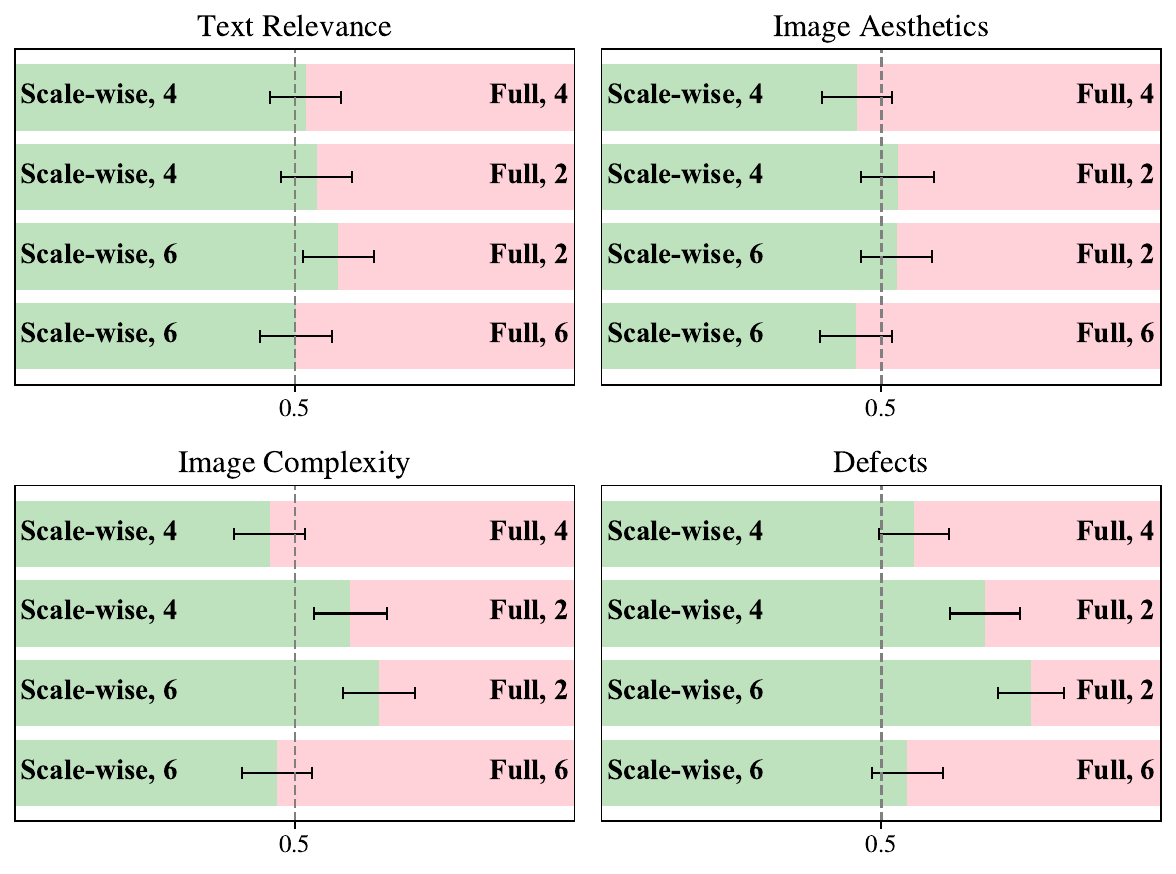}
\vspace{-4mm}
\caption{
    Visual examples (Left) and human preference study (Right) of the scale-wise and full-resolution settings within SD3.5 Medium. 
    The numbers indicate the sampling steps.
}
\label{fig:exps_full_vs_scale}
\end{figure}

We provide the quality comparisons for the SD3.5 Medium and evaluate FLUX in Appendix~\ref{app:full_vs_scale}.
Comparing the settings for the same number of steps ($4$ vs $4$, $6$ vs $6$), human evaluation (Figure~\ref{fig:exps_full_vs_scale}, Right) does not reveal any noticeable quality degradation.
Qualitative examples (Figure~\ref{fig:exps_full_vs_scale}, Left) further confirm this.
Interestingly, automatic metrics (Tables~\ref{tab:exps_full_vs_scale_coco} and~\ref{tab:exps_full_vs_scale_mjhq}) indicate that the scale-wise variants can even outperform their full-resolution counterparts, while being more efficient.

Then, we align generation times of scale-wise and full-resolution setups ($4$ vs. $2$, $6$ vs. $2$ steps) to assess quality differences. 
Human evaluation reveals a clear advantage for the scale-wise setup, particularly in reducing \textit{defects} and improving \textit{image complexity}. 
Examples in Figure~\ref{fig:exps_full_vs_scale} (Left) highlight the high defect rates of the $2$-step full-resolution baseline. Consistently, automatic metrics also show notable gains in HPSv3 and PS.

\textbf{Runtime.} 
Table~\ref{tab:exps_sampling_times} reports per-image generation latency (including VAE decoding and text encoding), and Table~\ref{tab:exps_train_times} shows average training iteration time. 
Compared to the full-resolution setting with the same number of steps ($4$ steps), the scale-wise setup achieves ${\sim}2{\times}$ speedup in both training and sampling across text-to-image models, and ${\sim}3{\times}$ for text-to-video.

\begin{table}
\begin{minipage}{0.47\textwidth}
\resizebox{\textwidth}{!}{%
\begin{tabular}{@{} l *{5}{c} @{}}
\toprule
\textbf{Setup} & \textbf{Steps} & \textbf{SD3.5-M} & \textbf{SD3.5-L} & \textbf{FLUX} & \textbf{Wan2.1} \\
\midrule
    Full-scale & $4$ & $0.29$ & $0.63$ & $1.41$ & $5.51$ \\
    Full-scale & $2$ & $0.16$ & $0.33$ & $0.72$ & $2.97$ \\
    Scale-wise & $6$ & $0.19$ & $0.41$ & $0.97$ & $2.61$ \\
    Scale-wise & $4$ & $0.17$ & $0.32$ & $0.72$ & $1.84$ \\
\bottomrule
\end{tabular}}
\vspace{-2mm}
\caption{
    Sampling times (sec / image) of scale-wise and full-resolution setups.
    The measurement setting is described in Appendix~\ref{app:measurement}.}
\label{tab:exps_sampling_times}
\end{minipage} \hfill
\begin{minipage}{0.51\textwidth}
\resizebox{\textwidth}{!}{%
\begin{tabular}{@{} l *{5}{c} @{}}
\toprule
\textbf{Setup} &  \textbf{Loss} & \textbf{SD3.5-M} & \textbf{SD3.5-L} &\textbf{FLUX} & \textbf{Wan2.1} \\
\midrule
    Full-scale & $\mathcal{L}_\text{SwD}$ & $7.5$ & $13.4$ & $22.8$ & $70.6$ \\
    Full-scale & $\mathcal{L}_\text{SwD{-}MMD}$ & $1.0$ & $1.7$ & $2.9$ & $12.7$ \\
    Scale-wise & $\mathcal{L}_\text{SwD}$ & $3.2$ & $7.8$ & $11.3$ & $23.9$ \\
    Scale-wise & $\mathcal{L}_\text{SwD{-}MMD}$ & $0.4$ & $0.9$ & $1.4$ & $4.4$ \\
\bottomrule
\end{tabular}}
\vspace{-2mm}
\caption{Training times (sec / iteration) for scale-wise and full-resolution $4$-step setups using the full objective ($\mathcal{L}_\text{SwD}$) and MMD only ($\mathcal{L}_\text{SwD-MMD}$).}
\label{tab:exps_train_times}
\end{minipage}
\end{table}


\subsection{Ablation study of MMD loss}
\label{sec:ablate_mmd}

\begin{figure}[t!]
\input{tables/ablation_mmd_mjhq}
\hfill
\begin{minipage}{0.48\textwidth}
    \centering
    \includegraphics[width=\linewidth]{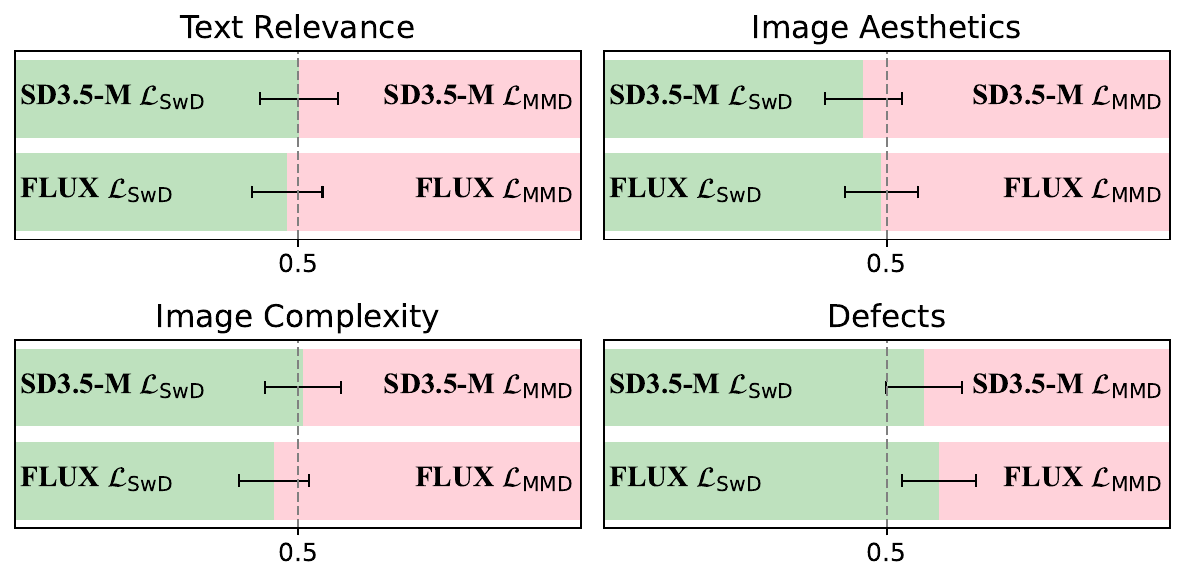}
    \vspace{-7mm}
    \caption{Comparison of the main SwD models against the ones distilled with $\mathcal{L}_\text{MMD}$ alone.}
    \label{fig:pdm_only_sbs}
\end{minipage}
\end{figure}

Here, we study the role of the $\mathcal{L}_\text{MMD}$ loss and its design choices. 
Most experiments are conducted with the $6$-step SD3.5-M setup from~\Cref{sec:main_results}, with the MJHQ results reported in Table~\ref{tab:exps_ablation_mmd_mjhq}.

We first assess the $\mathcal{L}_\text{MMD}$ contribution to $\mathcal{L}_\text{SwD}$. 
We observe that training with $\mathcal{L}_\text{MMD}$ alone slightly underperforms the full $\mathcal{L}_\text{SwD}$ but remains effective as an independent distillation method, whereas removing it from $\mathcal{L}_\text{SwD}$ leads to a significant drop in performance. 
Human evaluation (Figure~\ref{fig:pdm_only_sbs}) show that $\mathcal{L}_\text{MMD}$-only models exhibit noticeable degradation in defects, though not severe. 
Visual comparisons (Figure~\ref{fig:app_pdm_vs_swd}) confirm that they provide comparable performance to the full $\mathcal{L}_\text{SwD}$.
Moreover, as shown in Table~\ref{tab:exps_train_times}, $\mathcal{L}_\text{MMD}$-only training enables more than $7{\times}$ faster iterations since it avoids training extra models.

Finally, we examine several $\mathcal{L}_\text{MMD}$ variants.
The $\mathcal{L}_\text{MMD}$ with the RBF kernel (\textbf{A}) shows similar results.
Referring to the feature matching~\citep{salimans2016improved}, we consider two changes, \textbf{B}: the feature tokens in~\Cref{eq:mmd_loss} are averaged across the entire batch instead of per image, and \textbf{C}: extracting DM features only from clean samples, rather than noising them with the diffusion process.
We observe that both \textbf{B} and \textbf{C} make $\mathcal{L}_\text{MMD}$ less effective.

%% file: tables/main_table_video.tex
\begin{wraptable}{l}{0.46\textwidth}
\vspace{-2mm}
\resizebox{0.46\textwidth}{!}{%
\begin{tabular}{@{} l *{4}{c} @{}}
\toprule
\textbf{Model} & \begin{tabular}{@{}c@{}} \textbf{Latency,}\\\textbf{s/video} \end{tabular} &  \begin{tabular}{@{}c@{}} \textbf{Vision} \\ \textbf{Reward}  \end{tabular} $\uparrow$ & \begin{tabular}{@{}c@{}}\textbf{Video} \\ \textbf{Reward} \end{tabular} $\uparrow$ & \begin{tabular}{@{}c@{}}\textbf{VBench2} \\ \textbf{Overall}\end{tabular} $\uparrow$\\
\midrule
    Wan 2.1        & $137$ & $0.038$ & $5.43$ & $51.6$ \\
    CausVid$^*$      & $4.2$ & $0.042$ & $\mathbf{6.21}$ & $52.3$ \\
    \textbf{Spatial SwD}    & $2.1$ & $\mathbf{0.064}$ & $6.15$ & $52.8$ \\
    \textbf{SwD}           & $\mathbf{1.8}$ & $\mathbf{0.064}$ & $\mathbf{6.27}$ & $\mathbf{53.2}$ \\
\bottomrule
\addlinespace[1ex]
\multicolumn{5}{l}{\begin{tabular}{@{}l@{}} (*) The version trained on our Wan 2.1 generated dataset \\ \hspace{4mm} using the official implementation and setup.\end{tabular}} \\

\end{tabular}}
\vspace{-2mm}
\caption{
    Comparison of $4$-step \ourmethod variants with the Wan 2.1 and $3$-step CausVid models. 
}
\vspace{-4mm}
\label{tab:exps_main_results_video}
\end{wraptable}

%% file: tables/main_table_hpsv3.tex
\begin{table*}[t!]
\centering
\resizebox{1.0\linewidth}{!}{%
\begin{tabular}{lccccccccccccc}
\toprule
Model & Steps & \begin{tabular}{@{}c@{}}Latency,\\s/image \end{tabular} & \begin{tabular}{@{}c@{}}Model\\size, B\end{tabular} & PS $\uparrow$ & HPSv3 $\uparrow$ & IR $\uparrow$ & FID $\downarrow$ & PS $\uparrow$ & HPSv3 $\uparrow$ & IR $\uparrow$ & FID $\downarrow$ & GenEval
$\uparrow$ \\
& & & & \multicolumn{4}{c}{COCO $30$K} & \multicolumn{4}{c}{MJHQ $30$K} & \\
\cmidrule(lr){1-1}
\cmidrule(lr){2-4}
\cmidrule(lr){5-8}
\cmidrule(lr){9-12}
\cmidrule(lr){13-13}
Switti & $14$ & $0.44$ & $2.5$ & $22.6$ & $11.1$ & $0.98$ & $20.0$ & $21.6$ & $9.8$ & $0.84$ & $8.9$ & $0.62$ \\
Infinity & $14$ & $0.80$ & $2.0$ & ${22.7}$ & ${11.8}$ & $0.94$ & $28.1$ & $21.5$ & $10.5$ & $0.98$ & $12.9$ & $0.69$ \\
\cmidrule(lr){1-1}
\cmidrule(lr){2-4}
\cmidrule(lr){5-8}
\cmidrule(lr){9-12}
\cmidrule(lr){13-13}
SDXL & $50$ & $4.0$ & $2.6$ & $22.5$ & $9.4$ & $0.81$ & $\underline{14.8}$ & $\mathbf{21.6}$ & $9.2$ & ${0.83}$ & $\mathbf{8.1}$ & $
{0.56}$ \\
SDXL-Turbo & $4$ & $0.12^*$ & $2.6$ & ${22.6}$ & ${10.0}$ & $0.83$ & ${17.5}$ & $\underline{21.3}$ & $9.6$ & ${0.84}$ & $15.4$ & $0.55$ \\
Hyper-SD & $4$ & $\underline{0.20}$ & $2.6$ & $\underline{22.8}$ & $11.0$ & $\underline{0.90}$ & $20.1$ & $\mathbf{21.6}$ & $\underline{10.1}$ & $\underline{0.94}$ & $14.7$ & $0.55$ \\
SDXL-DMD2 & $4$ & $\underline{0.20}$ & $2.6$ & $\underline{22.8}$ & $\underline{12.0}$ & ${0.87}$ & $\mathbf{14.1}$ & $\mathbf{21.6}$ & $\underline{10.1}$ & ${0.86}$ & $\underline{8.3}$ & $\mathbf{0.58}$ \\
\textbf{SDXL-SwD} & $4$ & $\mathbf{0.11}$ & $2.6$ & $\mathbf{22.9}$ & $\mathbf{12.4}$ & $\mathbf{0.95}$ & $21.3$ & $\mathbf{21.6}$ & $\mathbf{10.3}$ & $\mathbf{0.97}$ & $15.1$ & $\underline{0.57}$ \\

\cmidrule(lr){1-1}
\cmidrule(lr){2-4}
\cmidrule(lr){5-8}
\cmidrule(lr){9-12}
\cmidrule(lr){13-13}
SD3.5-M  & $40$ & $4.8$ & $2.2$ & $\underline{22.4}$ & $\underline{10.2}$ & $\underline{1.00}$ & $\mathbf{16.3}$ & $\underline{21.6}$ & $\underline{9.9}$ & $\underline{0.97}$ & $\mathbf{9.5}$ & $\underline{0.69}$ \\
SD3.5-M-Turbo  & $4$ & $\underline{0.96}$ & $2.2$ & $22.2$ & $9.6$ & $0.83$ & $\underline{17.6}$ & $21.3$ & $9.3$ & $0.74$ & $13.6$ & $0.59$ \\
\textbf{SD3.5-M-SwD}  & $6$ & $\mathbf{0.19}$ & $2.2$ & $\mathbf{22.8}$ &  $\mathbf{11.7}$ & $\mathbf{1.12}$ & ${23.1}$  &$\mathbf{21.8}$ & $\mathbf{10.7}$ & $\mathbf{1.10}$ & $\underline{13.4}$ & $\mathbf{0.70}$ \\
\cmidrule(lr){1-1}
\cmidrule(lr){2-4}
\cmidrule(lr){5-8}
\cmidrule(lr){9-12}
\cmidrule(lr){13-13}
SD3.5-L  & $28$ & $8.3$ & $8.0$ & $\mathbf{22.8}$ & $\underline{11.3}$ & $\underline{1.06}$ & $\mathbf{16.5}$ & $\mathbf{21.8}$ & $\underline{10.4}$ & $\underline{1.04}$ & $\mathbf{10.7}$ & $\underline{0.70}$ \\
SD3.5-L-Turbo  & $4$ & $\underline{0.63}$ & $8.0$ & $\mathbf{22.8}$ & ${10}$ & $0.93$ & ${22.6}$ & $\underline{21.7}$ & $9.9$ & $0.9$ & $\underline{13.5}$ & $\underline{0.70}$ \\
\textbf{SD3.5-L-SwD}  & $4$ & $\mathbf{0.39}$ & $8.0$ & $\mathbf{22.8}$ & $\mathbf{12.8}$ & $\mathbf{1.20}$ & $\underline{20.6}$  & $\mathbf{21.8}$ & $\mathbf{11.1}$ & $\mathbf{1.22}$ & $13.9$ & $\mathbf{0.71}$ \\
\cmidrule(lr){1-1}
\cmidrule(lr){2-4}
\cmidrule(lr){5-8}
\cmidrule(lr){9-12}
\cmidrule(lr){13-13}
FLUX  & $30$ & $10.0$ & $12.0$ & $22.9$ & $12.4$ & $1.03$ & $23.6$ & $\underline{21.7}$ & $10.7$ & $0.93$ & $13.0$ & $0.66$ \\
FLUX-Turbo-Alpha  & $8$ & $2.75$ & $12.0$ & $\mathbf{23.1}$ & $\underline{13.4}$ & $\underline{1.08}$ & $\underline{21.2}$  & $21.5$ & $\underline{11.2}$ & $\underline{0.97}$ & $\underline{11.3}$ & $0.66$ \\
Hyper-FLUX  & $8$ &  $2.75$ & $12.0$ & $\underline{23.0}$ &  $12.4$ &  $0.94$  & $24.2$ & $\underline{21.7}$ &  $10.9$  & $0.85$ & $14.9$  &  $0.61$ \\
FLUX-Schnell & $4$ & $\underline{1.41}$ & $12.0$ & $22.6$ & $11.2$ & $1.01$ & $\mathbf{16.5}$  &$21.5$ & $10.3$ & $0.96$ & $\mathbf{9.8}$ & $\underline{0.69}$ \\
\textbf{FLUX-SwD}  & $4$ & $\mathbf{0.72}$ & $12.0$ & $\mathbf{23.1}$ &  $\mathbf{14.6}$ & $\mathbf{1.14}$ & $26.4$  & $\mathbf{21.9}$ & $\mathbf{11.6}$ & $\mathbf{1.06}$ & $14.4$ & $\mathbf{0.71}$ \\
\bottomrule
\addlinespace[1ex]
\multicolumn{13}{l}{(*) SDXL-Turbo is the only model that generates at $512{\times}512$, while all other models produce $1024{\times}1024$ images.} \\
\end{tabular}}
\vspace{-1mm}
\caption{
    Quantitative comparison of \ourmethod against other leading open-source models. 
    \textbf{Bold} indicates the best-performing model within each DM group, while \underline{underline} denotes the second best.
}
\label{tab:exps_main_results}
\end{table*}

%% file: tables/ablation_mmd_mjhq.tex
\begin{minipage}{0.47\textwidth}
\centering
\resizebox{\linewidth}{!}{
\begin{tabular}{@{} l *{5}{c} @{}}
\toprule
$\mathcal{L}_\text{SwD}$ setup  & PS $\uparrow$ & HPSv3 $\uparrow$ & IR $\uparrow$ & FID $\downarrow$ \\
\midrule
        $\mathcal{L}_\text{SwD}$ (Main) & $\mathbf{21.8}$ & $\mathbf{10.7}$ & $1.11$ & $\mathbf{13.6}$  \\
        \midrule
        $\mathcal{L}_\text{MMD}$ only & $21.5$ & $10.5$ & $\mathbf{1.15}$ & $13.8$  \\
        $\mathcal{L}_\text{SwD}$ w/o $\mathcal{L}_\text{MMD}$ & ${21.2}$ & $9.7$ & ${0.91}$ & $19.5$ \\
        \midrule
        \multicolumn{5}{c}{$\mathcal{L}_\text{MMD}$ Ablation} \\
        \midrule
        \textbf{A}: RBF kernel  & $\mathbf{21.8}$ & $\mathbf{10.8}$ & ${1.09}$ & $13.7$ \\
        \textbf{B}: Batch averaging & ${21.5}$ & ${10.5}$ & ${0.97}$ & $16.4$ \\
        \textbf{C}: w/o noising & ${21.3}$ & ${10.2}$ & ${1.01}$ & $16.6$ \\
        
\bottomrule
\end{tabular}}
\vspace{-3mm}
\captionof{table}{Ablation study of the $\mathcal{L}_\text{MMD}$ objective for SD3.5-Medium SwD on MJHQ30K.} \vspace{-3mm}
\label{tab:exps_ablation_mmd_mjhq}
\end{minipage}

%% file: sec/6_conclusion_arxiv.tex
\section{Conclusion}
We introduced SwD, a scale-wise diffusion distillation framework equipped with a novel patch-level MMD-based distillation technique.
We show that both components can be readily combined with existing state-of-the-art distillation methods and lead to further efficiency and quality improvements for few-step models.
We believe the proposed loss for DM distillation offers substantial potential for further development to pave the way toward a highly effective, self-contained distillation pipeline that eliminates the need for additional trainable models.




%% file: sec/7_supplementary.tex
\appendix
    



\section{Implementation details}
\label{app:implement_details}


Our few-step generator is initialized from a pretrained teacher DM, with trainable LoRA adapters~\citep{hu2022lora} added to the attention and MLP layers.
We use a LoRA rank of $64$ for SDXL, SD3.5-M, and SD3.5-L, and $128$ for FLUX and Wan 2.1. 
However, the rank has negligible impact on performance in our experiments.


\textbf{Training}. 
The model is trained to minimize $\mathcal{L}_{\text{SwD}} = \mathcal{L}_{\text{MMD}} + \alpha\cdot\mathcal{L}_{\text{DMD}} + \beta\cdot\mathcal{L}_{\text{GAN}}$. 

\textbf{$\mathcal{L}_\text{DMD}$} corresponds to the \textit{reverse KL divergence}, which employs the \textit{scores} of the ``real'' and ``fake'' distributions. 
The real score is estimated with the pretrained DM, while the fake score is modeled by training a separate ``fake'' DM on student-generated samples during distillation.
The fake model is parameterized with another set of LoRA adapters added to the teacher DM.
Real score estimation uses CFG~\citep{ho2022classifier} with corresponding default guidance scales for each model.
 
\textbf{$\mathcal{L}_\text{GAN}$}. 
Following DMD2~\citep{yin2024improved}, we also include a GAN loss for text-to-image settings. 
The discriminator is a $4$-layer MLP head operating on averaged spatial features extracted from the $11$-th (SD3.5-M), $20$-th (SD3.5-L), $15$-th (FLUX, Wan2.1) transformer blocks of the fake DM.
For SDXL, the feature maps are extracted from the middle UNet block.
The LoRA adapters of the fake DM are also updated using the discriminator loss.

\textbf{$\mathcal{L}_\text{MMD}$}.
For the proposed MMD loss, we use the timestep interval $[0, 600]$ ($[0, 400]$ for SDXL) to noise input samples prior to the feature extraction.
The transformer blocks for feature extraction are the same as those used in the GAN setting.

For the SD3.5-M, SD3.5-L, FLUX models, we use all losses in $\mathcal{L}_{\text{SwD}}$. 
For SDXL and Wan2.1, we use $\mathcal{L}_{\text{MMD}}$ only since $\mathcal{L}_{\text{DMD}}$ degrades the performance for these models as discussed in the following section, while $\mathcal{L}_{\text{GAN}}$ does not show noticeable improvements in our experiments. 

Overall, the models are trained with a learning rate of $4e{-}6$ and batch sizes of $64$ (SD3.5-M, SDXL) and $24$ (FLUX, SD3.5-L, Wan 2.1) for ${\sim}3K$ iterations on a single node with 8 A100 GPUs.

\textbf{Data.}
To exclude the effect of external data, we train the models on the teacher synthetic data, generated prior to distillation using the standard teacher setups.
SDXL: $50$ sampling steps with a guidance scale of $7.5$, followed by $10$ refiner steps.
SD3.5 Medium: $40$ sampling steps with a guidance scale of $4.5$. 
SD3.5 Large: $28$ sampling steps with a guidance scale of $4.5$.
FLUX: $30$ sampling steps with a guidance scale of $4.5$.
Wan2.1: $50$ steps with a guidance scale of $5.0$.

\section{Importance of Scale-adapted Teacher Models}
\label{app:scale_adapted_teacher}

\begin{wrapfigure}{l}{0.5\textwidth}
    \centering
    \vspace{-4mm}
    \includegraphics[width=\linewidth]{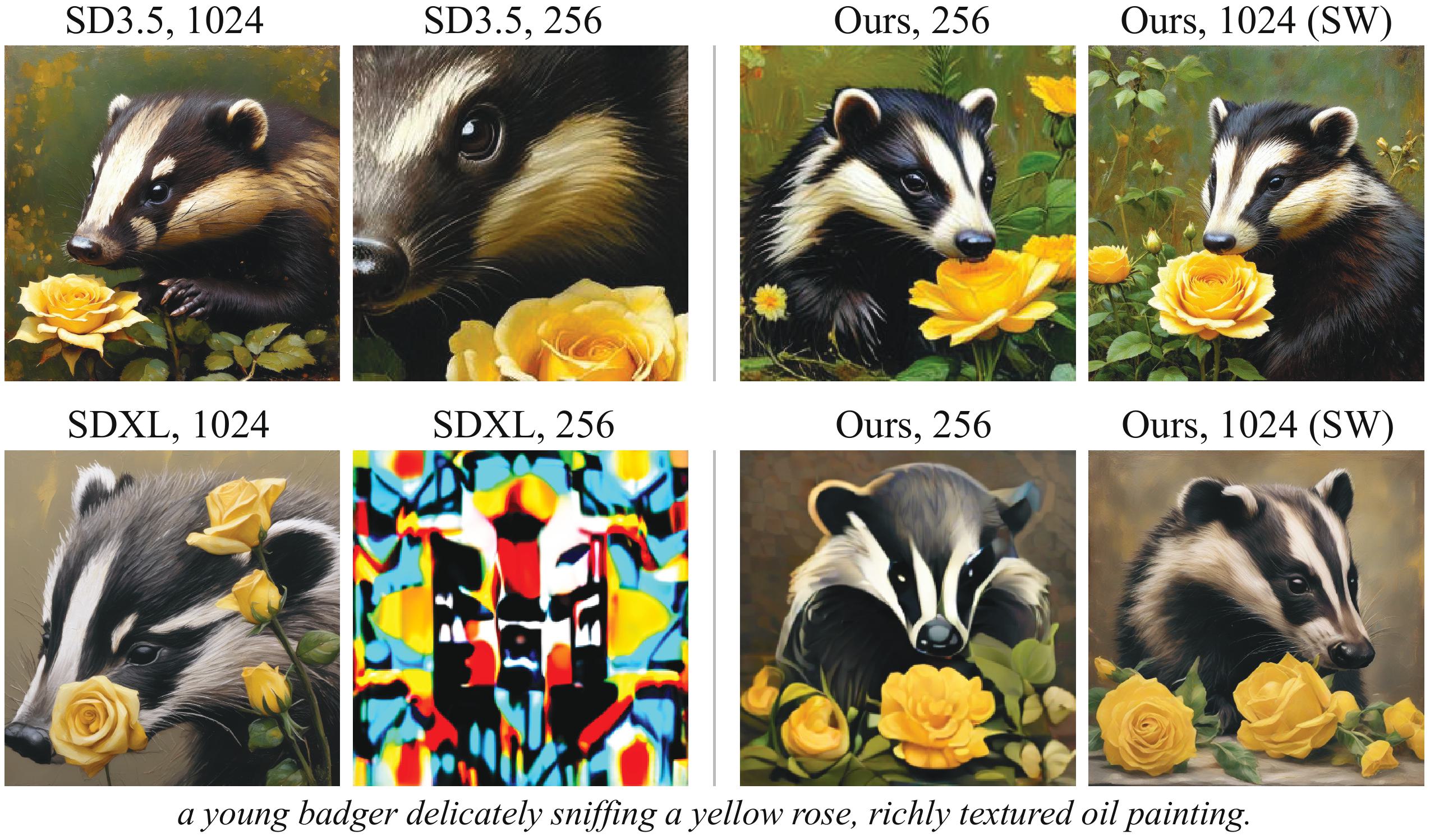}
    \vspace{-6mm}
    \caption{
        SD3.5 generates cropped images at low-resolutions ($256{\times}256$), while SDXL does not produce meaningful images at all.
        \ourmethod successfully adapts the model to both fixed low-resolution sampling (Ours, 256) and scale-wise sampling (Ours, 1024 SW) thanks to $\mathcal{L}_\text{GAN}$ and $\mathcal{L}_\text{MMD}$.
        In contrast, $\mathcal{L}_\text{DMD}$ tends to inherit teacher's limitations and thereby should be applied with caution or disabled altogether in such cases.
    }
    \vspace{-5mm}
    \label{fig:exps_crops}
\end{wrapfigure}

We also address the following important question: 
\textit{Does the teacher model need to be capable of generating images at low scales prior to scale-wise distillation?} 
The teacher model may not inherently handle low scales effectively, making scale-wise distillation more challenging than the full-scale distillation. 
If this limitation is significant, additional pretraining of the teacher on small scales might be required.

We evaluate the ability of SD3.5 Medium and SDXL to generate images at a lower resolution ($256{\times}256$).
The results are shown in Figure~\ref{fig:exps_crops}.

We observe that SD3.5 produces cropped and simplified images, but the overall quality remains acceptable, likely due to its pretraining at $256{\times}256$ resolution. 
\ourmethod successfully distills this model, mitigating the cropping problem and increasing image complexity in both the fixed low-resolution setting (Ours, 256) and under scale-wise sampling (Ours, 1024 SW).

More importantly, SDXL fails to generate faithful images at $256{\times}256$ resolution. 
Interestingly, \ourmethod is able to recover from such a poor starting point during distillation.

We attribute \ourmethod’s ability to adapt to lower-resolution sampling to the $\mathcal{L}_\text{GAN}$ and $\mathcal{L}_\text{MMD}$ objectives that use low-resolution target images during distillation.
In contrast, $\mathcal{L}_\text{DMD}$ does not rely on reference images at the target resolution. 
Empirically, we find that $\mathcal{L}_\text{DMD}$ degrades student performance when applied to teachers that struggle with low-resolution generation (e.g., SDXL and Wan2.1).

In summary, \textbf{$\mathcal{L}_\text{DMD}$ is sensitive to the teacher's ability to generate in low resolution and tends to inherit its limitations, whereas $\mathcal{L}_\text{GAN}$ and $\mathcal{L}_\text{MMD}$ are significantly more robust.}

Therefore, in our main experiments, we perform scale-wise distillation for the SDXL and Wan 2.1 models using $\mathcal{L}_\text{MMD}$ only.
We excluded $\mathcal{L}_\text{GAN}$ for these models as it did not show significant additional improvements.

\begin{table}
\vspace{-2mm}
\input{tables/full_vs_scale_coco}
\input{tables/full_vs_scale_mjhq}
\end{table}

\begin{figure}[t!]
\begin{minipage}{0.47\textwidth}
    \centering
    \vspace{-2mm}
    \includegraphics[width=1.0\linewidth]{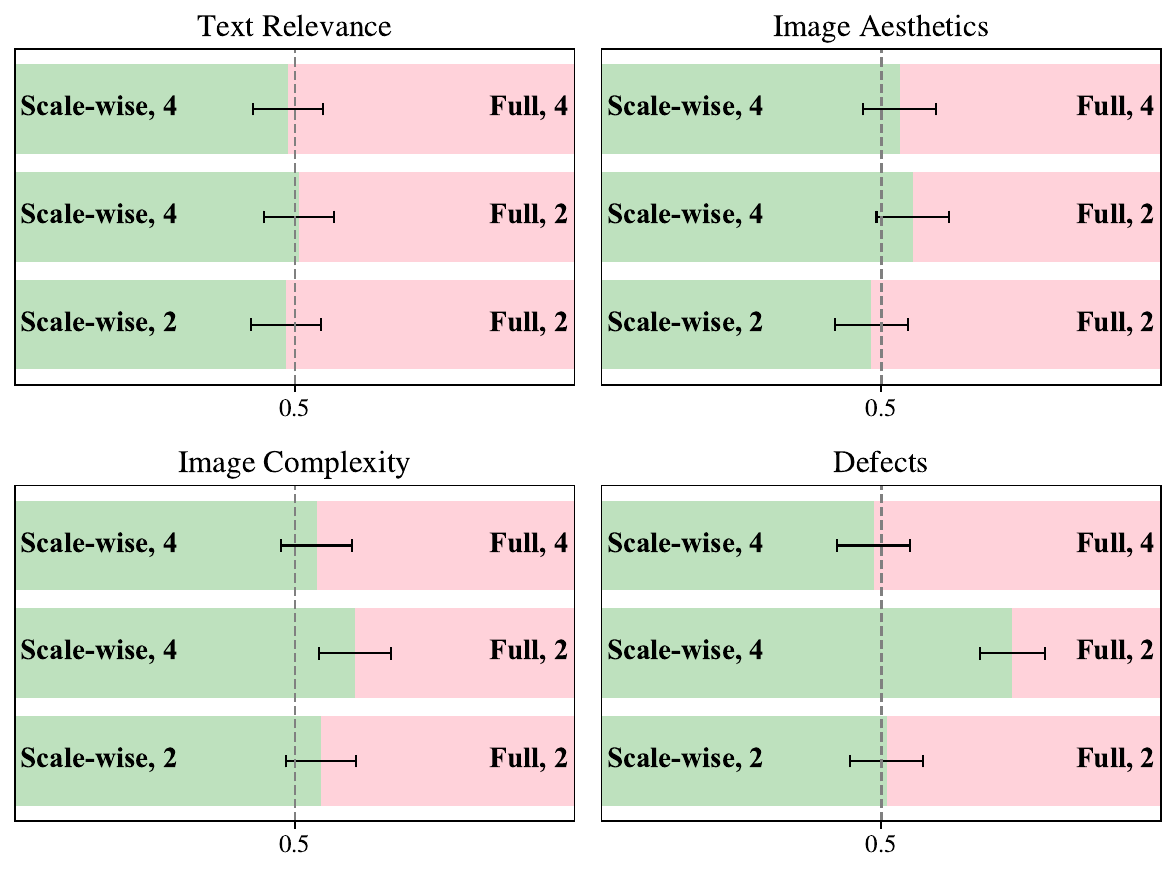}
    \vspace{-7mm}
    \caption{Human preference study comparing scale-wise and full-resolution FLUX setups.}
\label{fig:app_full_vs_scale_flux}
\end{minipage}
\hfill
\begin{minipage}{0.50\textwidth}
\vspace{-2mm}
\centering
\resizebox{1.02\linewidth}{!}{%
\begin{tabular}{@{} l *{5}{c} @{}}
\toprule
Setup  & PS $\uparrow$ & HPSv3 $\uparrow$ & IR $\uparrow$ & FID $\downarrow$ \\
\midrule
\multicolumn{5}{c}{COCO2014} \\
\midrule
        Main $s{=}[32,48,64,80,96,128]$ & $\mathbf{22.8}$ & $\mathbf{11.7}$ & $\mathbf{1.10}$ & $\mathbf{23.1}$  \\
        $s{=}[64,64,64,64,64,128]$ & ${22.4}$ & $10.3$ & ${1.02}$ & $23.7$ \\
        $s{=}[32,32,32,32,32,128]$ & ${22.3}$ & $9.8$ & ${0.97}$ & $23.8$ \\
\midrule
\multicolumn{5}{c}{MJHQ} \\
\midrule
        Main $s{=}[32,48,64,80,96,128]$ & $\mathbf{21.8}$ & $\mathbf{10.7}$ & $\mathbf{1.11}$ & $\mathbf{13.6}$  \\
        $s{=}[64,64,64,64,64,128]$ & ${21.3}$ & $9.8$ & ${1.06}$ & $14.6$ \\
        $s{=}[32,32,32,32,32,128]$ & ${21.2}$ & $9.4$ & ${0.99}$ & $15.7$ \\
\bottomrule
\end{tabular}}
\vspace{-2mm}
\captionof{table}{Comparisons to the ``constant'' scale schedules for SD3.5-Medium SwD.} \vspace{-3mm}
\label{tab:app_constant_vs_main}
\vspace{-4mm}
\end{minipage}
\end{figure}

\section{SwD timestep and scale schedules}
\label{app:main_setups}

Below, we provide the timestep $t$ and scale $s$ schedules used in our main experiments.
The scale schedule shows the resolutions in the corresponding VAE latent spaces.

\textbf{SDXL.} 
$t=[1000,800,600,400]$, $s=[64,80,96,128]$.

\textbf{SD3.5 Medium.} $t=[1000,945,896,790,737,602]$, $s=[32,48,64,80,96,128]$.

\textbf{SD3.5 Large.} $t=[1000,896,737,602]$, $s=[64,80,96,128]$.

\textbf{FLUX.} $t=[1000,945,790,602]$, $s=[32,64,96,128]$.

\textbf{Wan2.1.} $t=[1000,896,737,602]$, $s=[6{\times}20{\times}34,11{\times}30{\times}52, 16{\times}40{\times}70,21{\times}60{\times}104]$.



\section{Scale-wise versus Full-resolution}
\label{app:full_vs_scale}

\begin{figure}[t!]
\vspace{2mm}
\begin{minipage}{0.49\textwidth}
    \centering
    \includegraphics[width=1.0\linewidth]{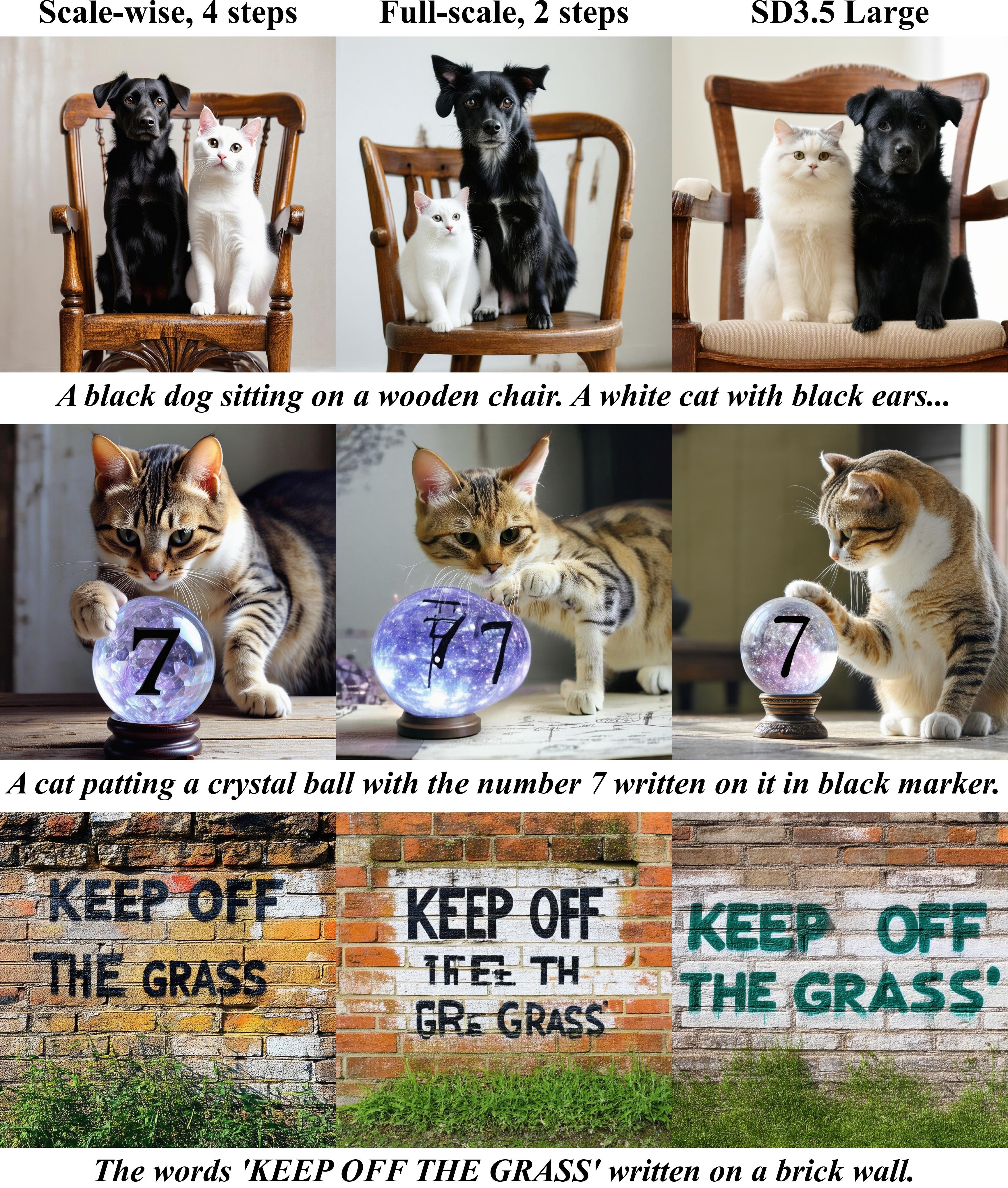}
    \vspace{-4mm}
    \caption{Visual examples of $4$-step scale-wise and $2$-step full-resolution SD3.5-Large settings.}
    \label{fig:app_large}
\end{minipage}
\hfill
\begin{minipage}{0.49\textwidth}
    \centering
    \includegraphics[width=1.0\linewidth]{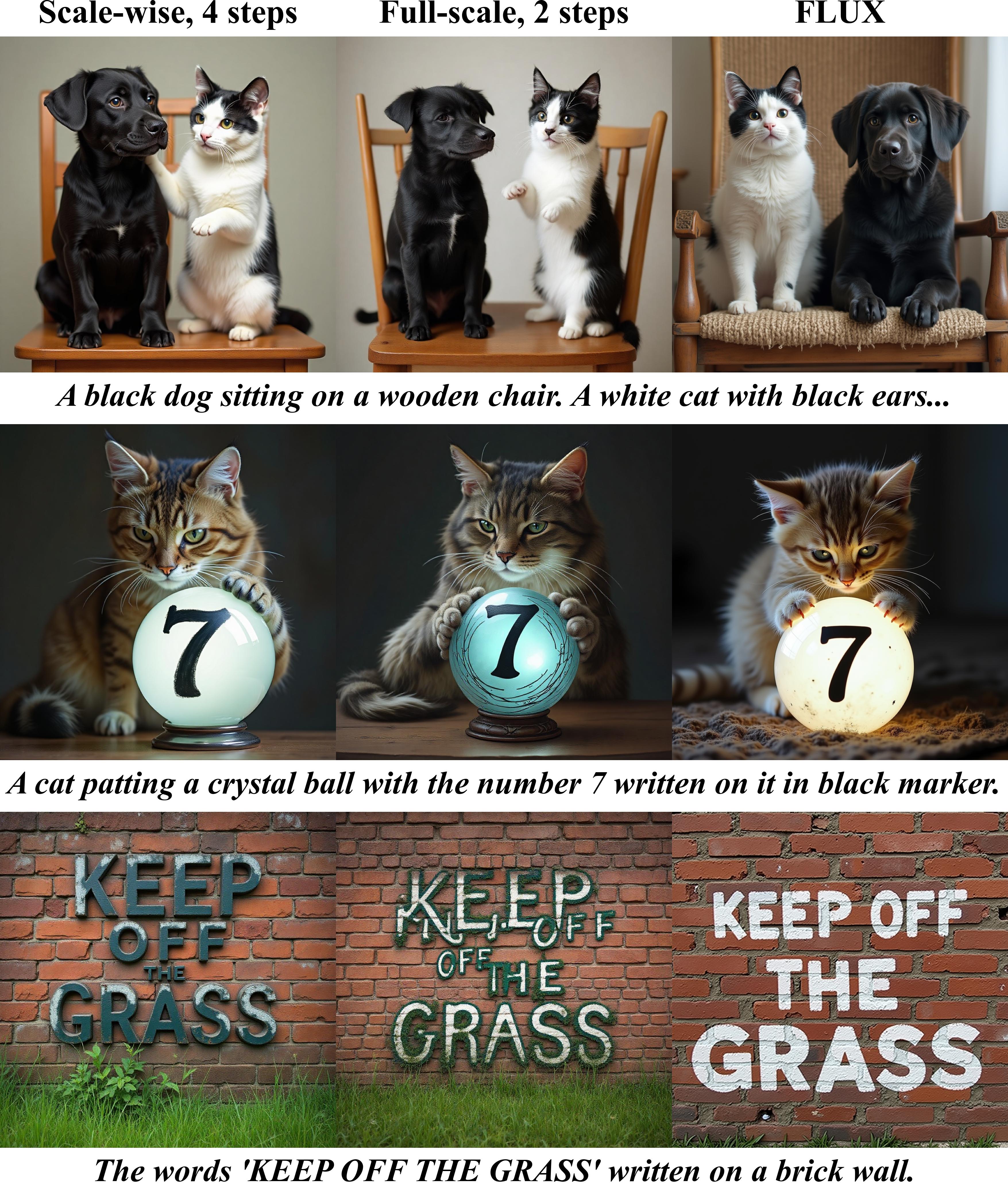}
    \vspace{-4mm}
    \caption{Visual examples of $4$-step scale-wise and $2$-step full-resolution FLUX settings.}
    \label{fig:app_flux}
\end{minipage}
\end{figure}

\begin{figure*}
    \centering
    \includegraphics[width=1.0\linewidth]{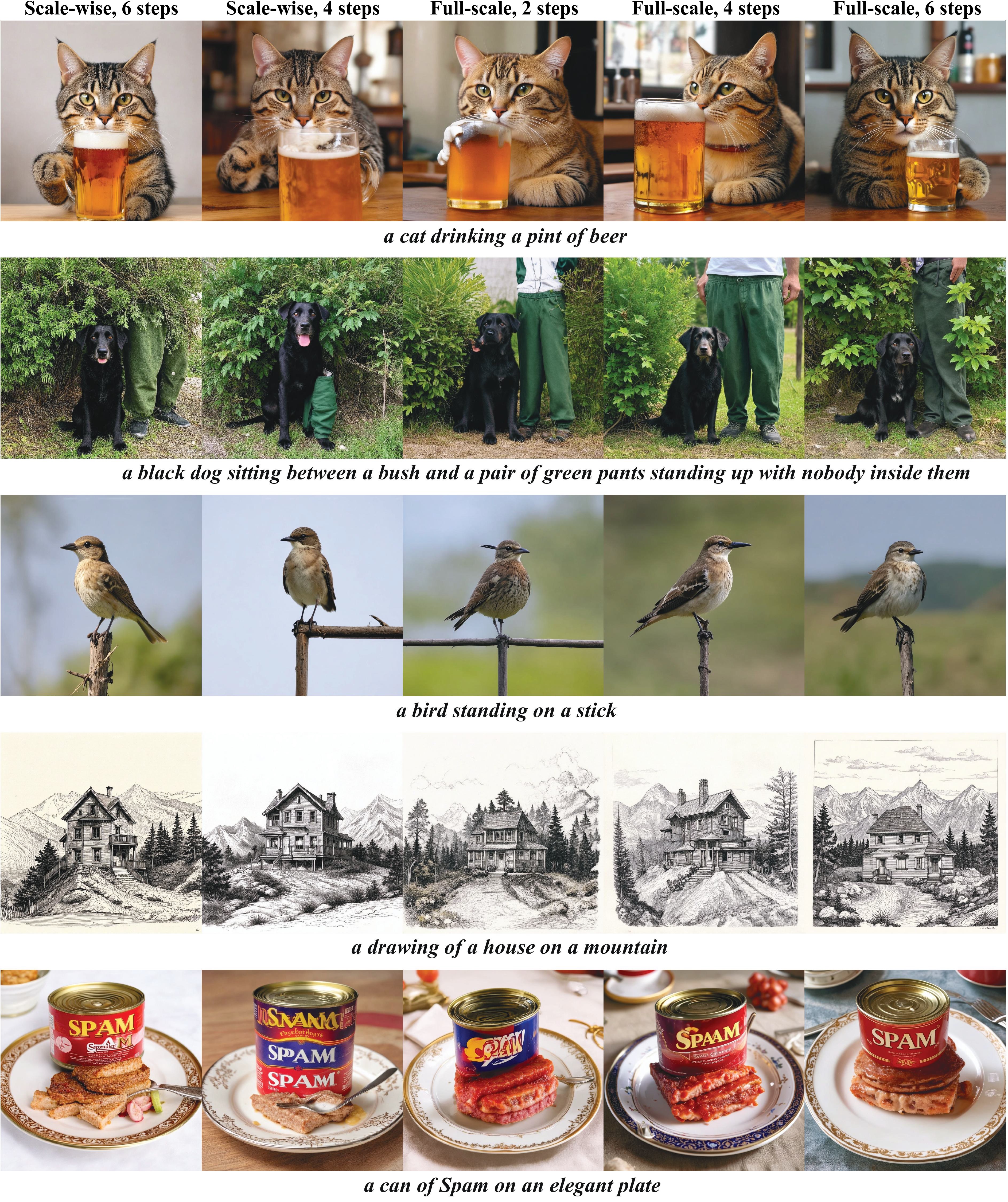}
    \caption{Qualitative examples of image generations using scale-wise and
full-resolution SD3.5 Medium SwD variants for different generation steps.}
    \label{fig:app_full_vs_scale_imgs}
\end{figure*}

Here, we provide additional results to compare the various scale-wise and full-resolution settings.
~\Cref{tab:exps_full_vs_scale_coco} and
\Cref{tab:exps_full_vs_scale_mjhq} present the automatic metrics for SD3.5 Medium and FLUX on the COCO and MJHQ datasets.
The visual examples for FLUX, SD3.5 Large, SD3.5 Medium are presented in Figures~\ref{fig:app_large}, ~\ref{fig:app_flux}, and ~\ref{fig:app_full_vs_scale_imgs}, respectively.

In \Cref{tab:app_constant_vs_main}, we evaluate the use of ``constant'' $6$-step scale schedules $s{=}[64,64,64,64,64,128]$ and $s{=}[32,32,32,32,32,128]$ for SD3.5 Medium in contrast to the progressively growing schedule $s{=}[32,48,64,80,96,128]$, used in our main setup.
Note that the last step is required to be made at the target resolution.
The results show that it is important to gradually increase the resolution over sampling steps.

\begin{wraptable}{l}{0.5\linewidth}
\centering
\vspace{-4mm}
\resizebox{\linewidth}{!}{%
\begin{tabular}{lccccc}
\toprule
& & \multicolumn{2}{c}{Patch FID $\downarrow$} & \multicolumn{2}{c}{Patch CLIP-IQA $\uparrow$}
\\
\midrule
Model & Steps & SwD & Full-scale & SwD & Full-scale \\
\midrule
SD3.5-M & $6$ & $20.1$ & $21.1$ & $0.85$ & $0.85$\\
SD3.5-L & $4$ & $20.4$ & $21.9$ & $0.88$ & $0.88$\\
FLUX    & $4$ & $18.7$ & $18.9$ & $0.80$ & $0.79$ \\
\bottomrule
\end{tabular}}
\vspace{-2mm}
\caption{Comparison of SwD and full-resolution distilled models in high-frequency detail preservation on MJHQ30K.}
\vspace{-3mm}
\label{tab:patch_fid}
\end{wraptable}


Finally, we investigate if scale-wise distillation affects high-frequency details compared to the models distilled at the target resolution. 
We evaluate Patch FID~\citep{lin2024sdxllightning} and Patch CLIP-IQA~\citep{wang2022exploring} on MJHQ30K.
As shown in~\Cref{tab:patch_fid}, SwD achieves similar scores to the full-resolution distilled models, indicating no noticeable degradation of high-frequency details.

\section{Extended latent space spectral analysis}
\label{app:extended_anal}
\begin{figure*}[t!]
    \centering
    \includegraphics[width=1.0\linewidth]{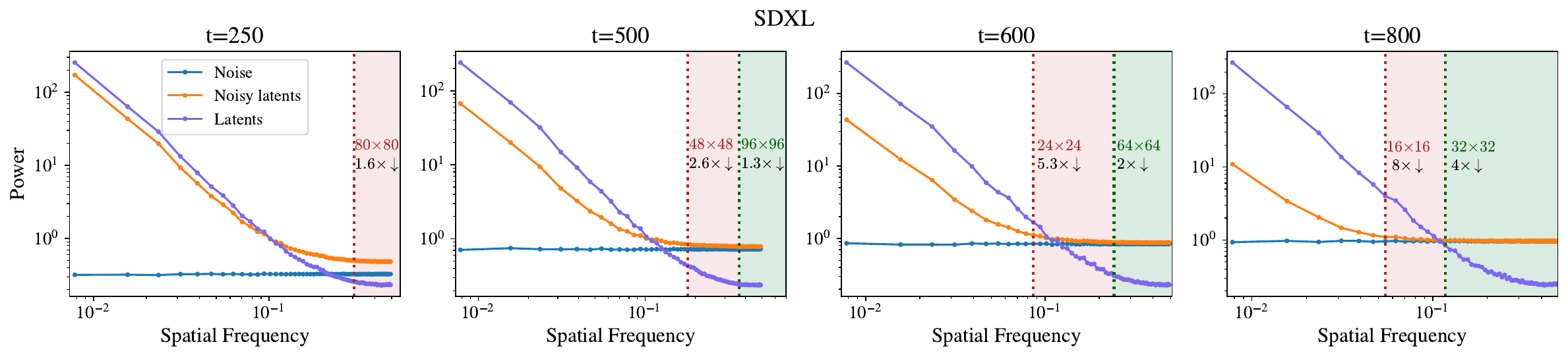}
    \includegraphics[width=1.0\linewidth]{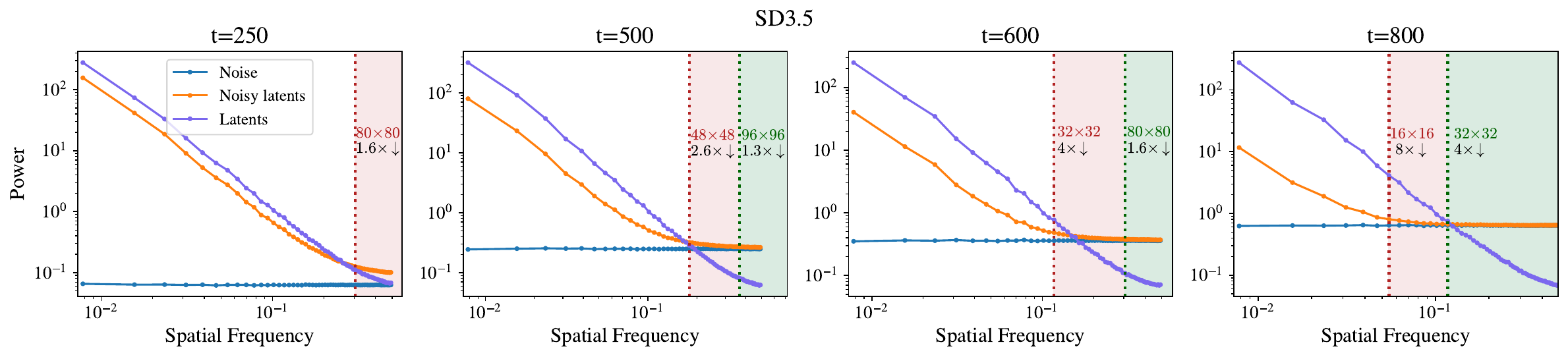}
    \includegraphics[width=1.0\linewidth]{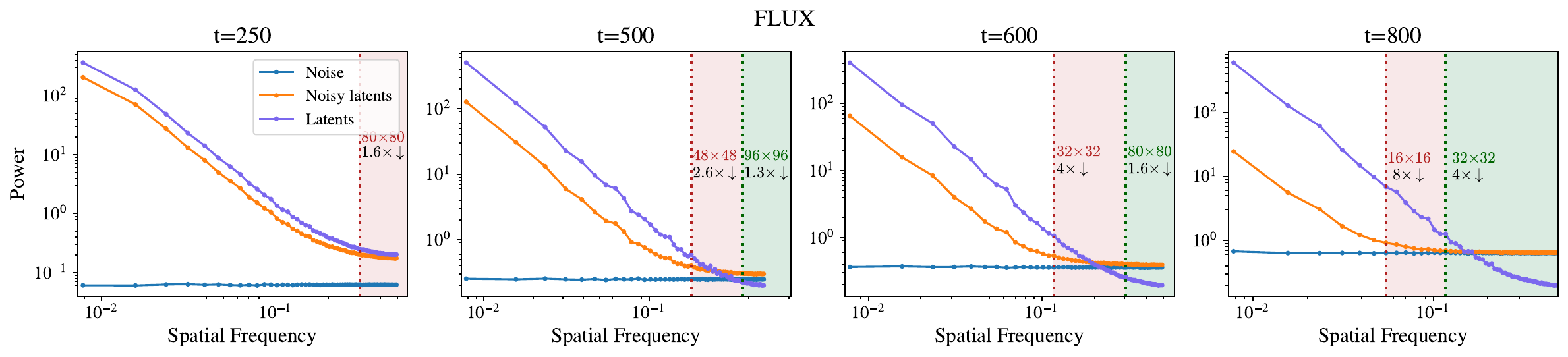}
    \includegraphics[width=1.0\linewidth]{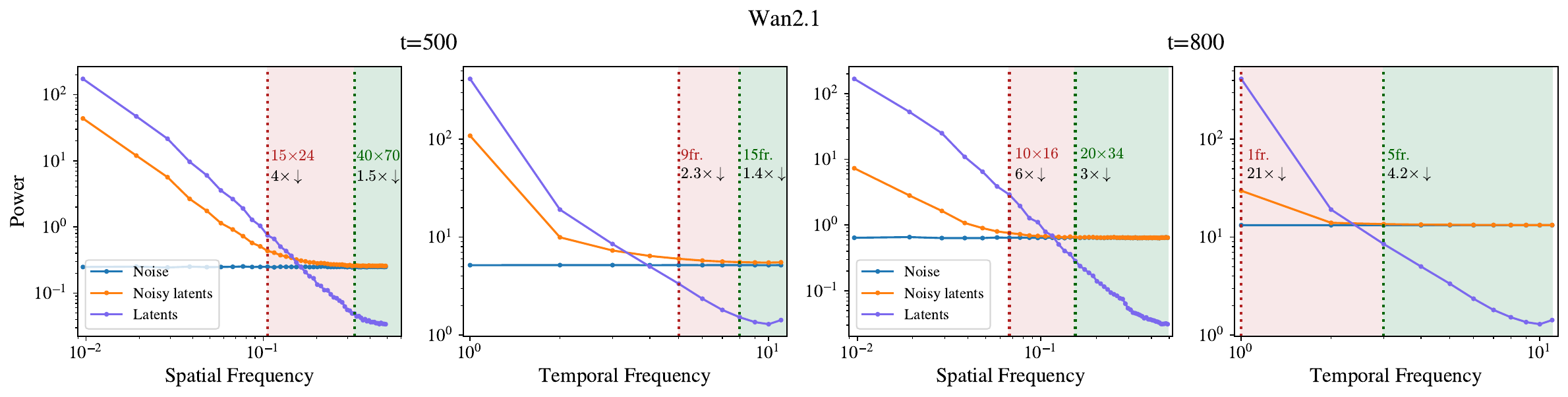}
    \caption{Extended spectral analysis from~\Cref{sec:analysis} to more timesteps and models (FLUX, SDXL).}
    \label{fig:anal_app}
\end{figure*}

We provide additional results for our latent space spectral analysis in~\Cref{sec:analysis}.
\Cref{fig:anal_app} provides more results for the SD3.5 and Wan2.1 models and also includes the analysis for FLUX~\citep{flux} and SDXL~\citep{podell2024sdxl}. 
In contrast to other models, the SDXL model~\citep{podell2024sdxl} uses a variance-preserving (VP) diffusion process~\citep{ho2020denoising,song2020score}.
The SDXL latent space has $C{=}4$ channels and $128{\times}128$ spatial resolution.

\section{Extended analysis of noisy latent upscaling strategies}
\label{app:extended_upscaling_strategies}
\input{tables/extended_table1}

In~\Cref{tab:extended_table1}, we present the extended comparison of noisy latent upscaling strategies for $32\times32 \rightarrow 128\times128$ and $96\times96 \rightarrow 128\times128$ setups.
Specifically, given a full-resolution ($128{\times}128$) real image latent, $\mathbf{{x}}_{0}$, and its downscaled versions $\mathbf{{x}}^{32{\times}32}_{0}$, $\mathbf{{x}}^{64{\times}64}_{0}$ and $\mathbf{{x}}^{128{\times}128}_{0}$.
The images are generated with Stable Diffusion 3.5 Medium~\citep{esser2024scaling} from intermediate noisy latents, $\mathbf{{x}}_{t}$, obtained with different upscaling strategies.
We use a default guidance scale of $7$ and the standard $28$-step timestep schedule.

Upscaling the noised low-resoluton latent provides poor performance across all upscaling factors. 
For upscaling $\mathbf{x}_0$-predictions, as expected, lower upscale ratios consistently lead to better results across all timesteps, since the predictions are less distorted by upscaling.
Also, we note that even for the $96{\times}96$ setup, the performance is still not fully recovered at $t{=}600$, suggesting that the upsampled latents are still slightly shifted and require the scale-wise model training.



\section{Runtime measurement setup}
\label{app:measurement}
In our experiments, we measure runtimes in half-precision (FP16), using \textit{torch.compile} for all models: VAE decoders, text encoders, and generators.
Note that, under our very fast sampling settings, the computational costs of the text encoder and VAE start to account for a noticeable portion of the overall runtime.
Thus, we replace original VAEs with TinyVAEs~\citep{BoerBohan2025TAEHV} for all models.

The measurements are conducted in an isolated environment on a single A100 GPU.
We use a batch size of $8$ for all runs, and each measurement is averaged over $100$ independent runs.
The latency is then obtained by dividing the average runtime by the batch size.

\section{Human evaluation}
\label{app:human_eval}
The evaluation is conducted using Side-by-Side (SbS) comparisons, where assessors are presented with two images alongside a textual prompt and asked to choose the preferred one. For each pair, three independent responses are collected, and the final decision is determined through majority voting.

The human evaluation is carried out by professional assessors who are formally hired, compensated with competitive salaries, and fully informed about potential risks. Each assessor undergoes detailed training and testing, including fine-grained instructions for every evaluation aspect, before participating in the main tasks.

\begin{figure}[t!]
    \centering
    \includegraphics[width=1.0\linewidth]{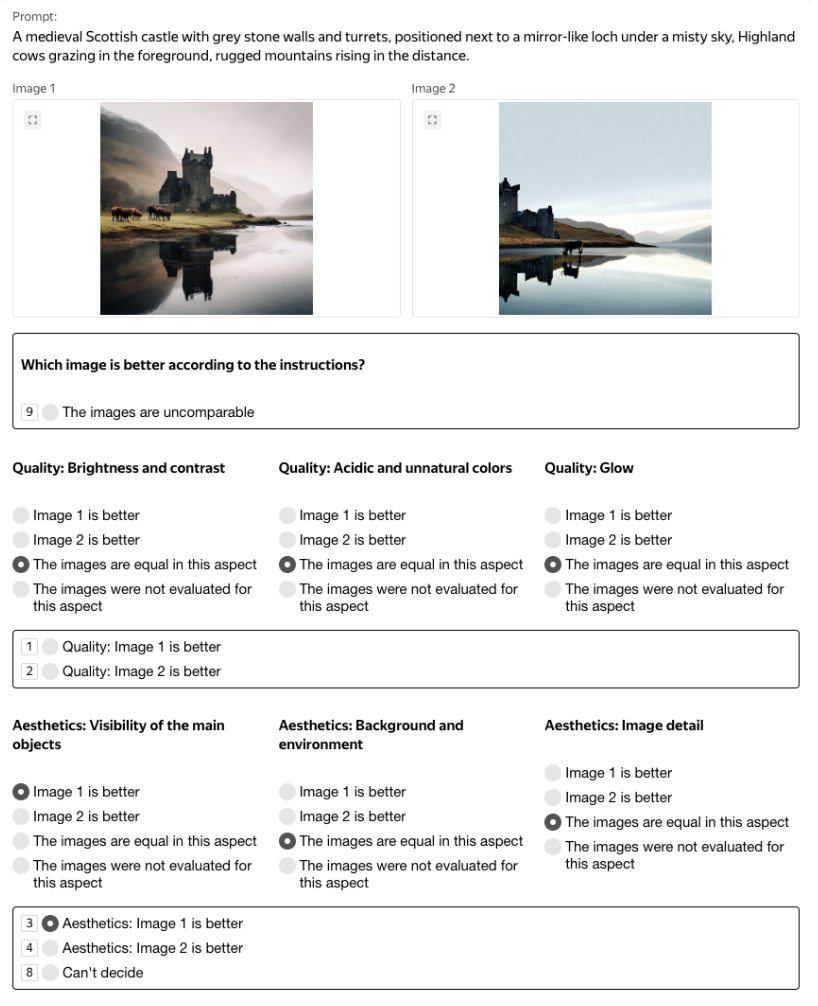}
    \caption{Human evaluation interface for aesthetics.}
    \label{fig:app_aesthetics}
\end{figure}

\begin{figure}[t!]
    \centering
    \includegraphics[width=1.0\linewidth]{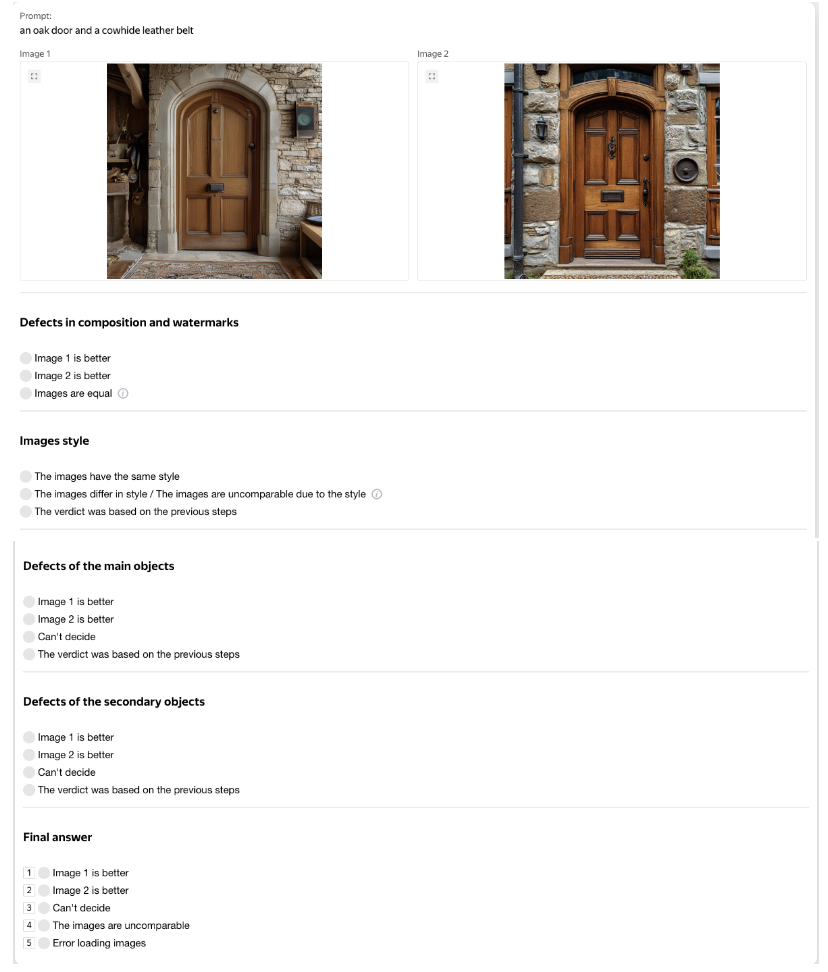}
    \caption{Human evaluation interface for defects.}
    \label{fig:app_defects}
\end{figure}

\begin{figure}[t!]
    \centering
    \includegraphics[width=1.0\linewidth]{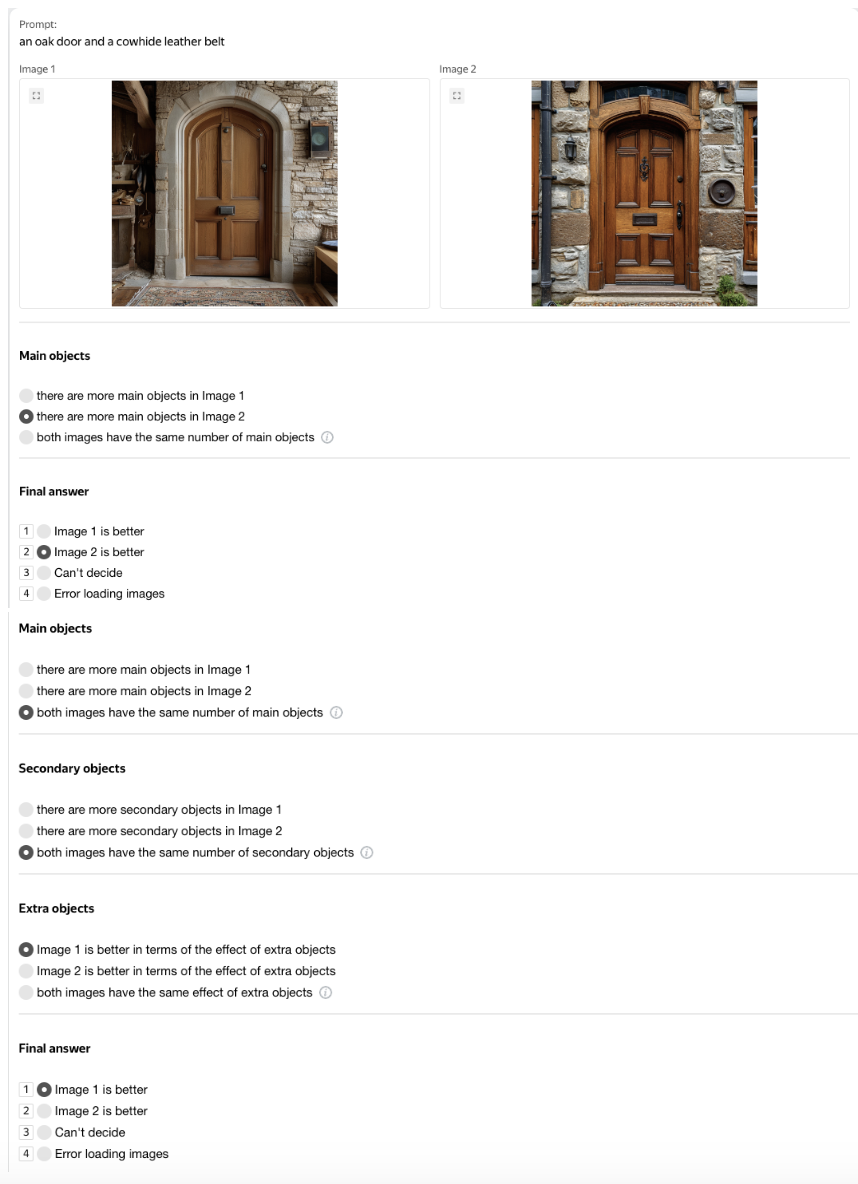}
    \caption{Human evaluation interface for relevance.}
    \label{fig:app_relevance}
\end{figure}

\begin{figure}[t!]
    \centering
    \includegraphics[width=1.0\linewidth]{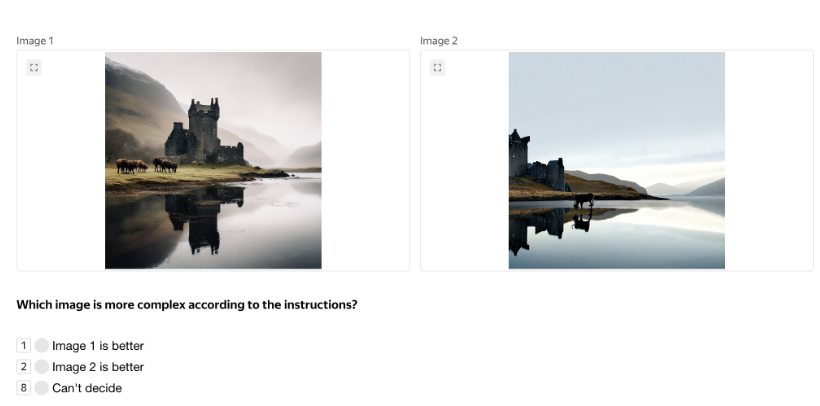}
    \caption{Human evaluation interface for complexity.}
    \label{fig:app_complexity}
\end{figure}

In our human preference study, we compare the models across four key criteria: relevance to the textual prompt, presence of defects, image aesthetics, and image complexity. Figures~\ref{fig:app_aesthetics}, \ref{fig:app_complexity}, \ref{fig:app_defects}, \ref{fig:app_relevance} illustrate the interface used for each criterion. Note that the images displayed in the figures are randomly selected for demonstration purposes.

\clearpage
\begin{table}[ht!]
    \centering
    \scriptsize
    \small
    \setlength{\tabcolsep}{3pt}
    \begin{adjustbox}{width=\textwidth}
    \begin{tabular}{l*{24}{c}}
        \toprule
        Method & Total & Creativity & Commonsense & Controllability & Human Fidelity & Physics & Human & Human  & Human  & Composition & Diversity & Mechanics & Material & Thermotics & Multi-View & Dynamic Spatial & Dynamic  & Motion Order  & Human  & Complex  & Complex  & Camera  & Motion  & Instance  \\  &  Score &  Score &  Score &  Score &  Score & Score &  Anatomy &  Clothes &  Identity &  &  &  &  &  &  Consistency & Relationship &  Attribute &   Understanding &  Interaction &  Landscape &  Plot &  Motion &  Rationality &  Preservation \\
        \midrule
        Wan 2.1 & $51.59$ & $\mathbf{53.75}$ & $57.06$ & $22.65$ & $83.03$ & $41.45$ & $87.00$ & $91.24$ & $70.85$ & $42.56$ & $\mathbf{64.94}$ & $\mathbf{59.16}$ & $36.58$ & $57.89$ & $12.15$ & $26.08$ & $15.01$ & \underline{$21.21$} & $46.33$ & $\mathbf{19.77}$ & $\mathbf{11.02}$ & \underline{$19.13$} & $28.16$ & \underline{$85.96$}\\
        CausVid & $52.31$ & {$39.94$} & $59.69$ & $21.24$ & $\mathbf{92.54}$ & $\mathbf{48.16}$ & $\mathbf{91.51}$ & \underline{$97.73$} & $\mathbf{88.40}$ & $46.47$ & {$33.41$} & $53.15$ & $37.33$ & \underline{$63.23$} & $\mathbf{38.95}$ & $26.57$ & \underline{$17.58$} & $18.51$ & $53.66$ & {$18.22$} & {$7.75$} & {$6.41$} & $27.58$ & $\mathbf{91.81}$\\
        \textbf{Spatial SwD} & \underline{$52.43$} &	\underline{$44.54$} &	\underline{$60.84$} &	$\mathbf{29.11}$ &	{$84.10$} &	$43.57$ &	$84.52$ &	{$94.54$} & {$73.26$} &	$\mathbf{53.05}$ & \underline{$36.03$} &	\underline{$58.33$} &	\underline{$44.68$} &	$56.83$ &	$14.45$ &	$\mathbf{35.74}$ &	$\mathbf{27.83}$ &	$\mathbf{21.88}$ &	$\mathbf{71.66}$ &	\underline{$19.11$} &	\underline{$10.59$} &	$16.97$ &	\underline{$29.88$} &	$\mathbf{91.81}$ \\
        \textbf{SwD} & $\mathbf{53.22}$ &	$40.44$ &	$\mathbf{63.72}$ &	\underline{$26.80$} &	\underline{$88.42$} &	\underline{$46.73$}&	\underline{$89.00$}&	$\mathbf{99.54}$ &	\underline{$76.73$}&	\underline{$47.30$} &	$33.59$ &	$57.14$ &	$\mathbf{47.36}$&	$\mathbf{63.5}$&	\underline{$18.94$}&	\underline{$33.33$}&	\underline{$17.58$}&	$20.20$&	\underline{$71.00$}&	$14.44$&	$10.07$&	$\mathbf{20.98}$ &	$\mathbf{35.63}$&	$\mathbf{91.81}$ \\
        \bottomrule
    \end{tabular}
    \end{adjustbox}
    \caption{Full comparison on VBench2.0 using all 18 metrics.}
\end{table}

\begin{figure*}[t!]
    \centering
    \vspace{-2mm}
    \includegraphics[width=0.85\linewidth]{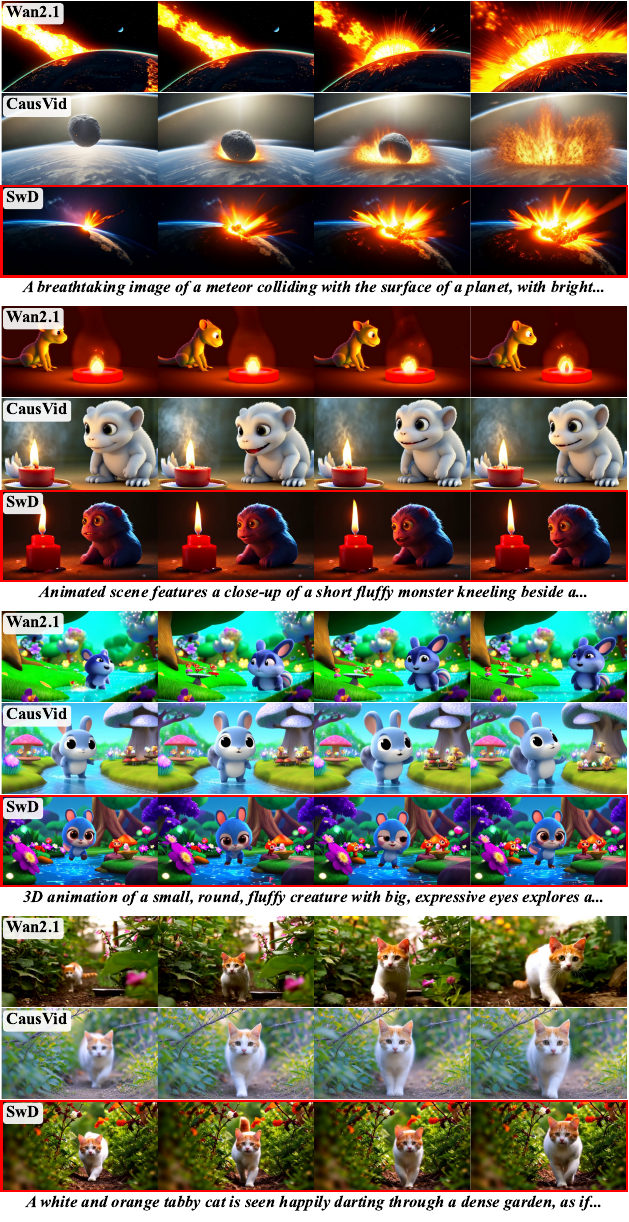}
    \vspace{-4mm}
    \caption{Qualitative results of Wan2.1-SwD.}
    \label{fig:app_cherry_video}
\end{figure*}

\begin{figure*}[t!]
    \centering
    \includegraphics[width=0.98\linewidth]{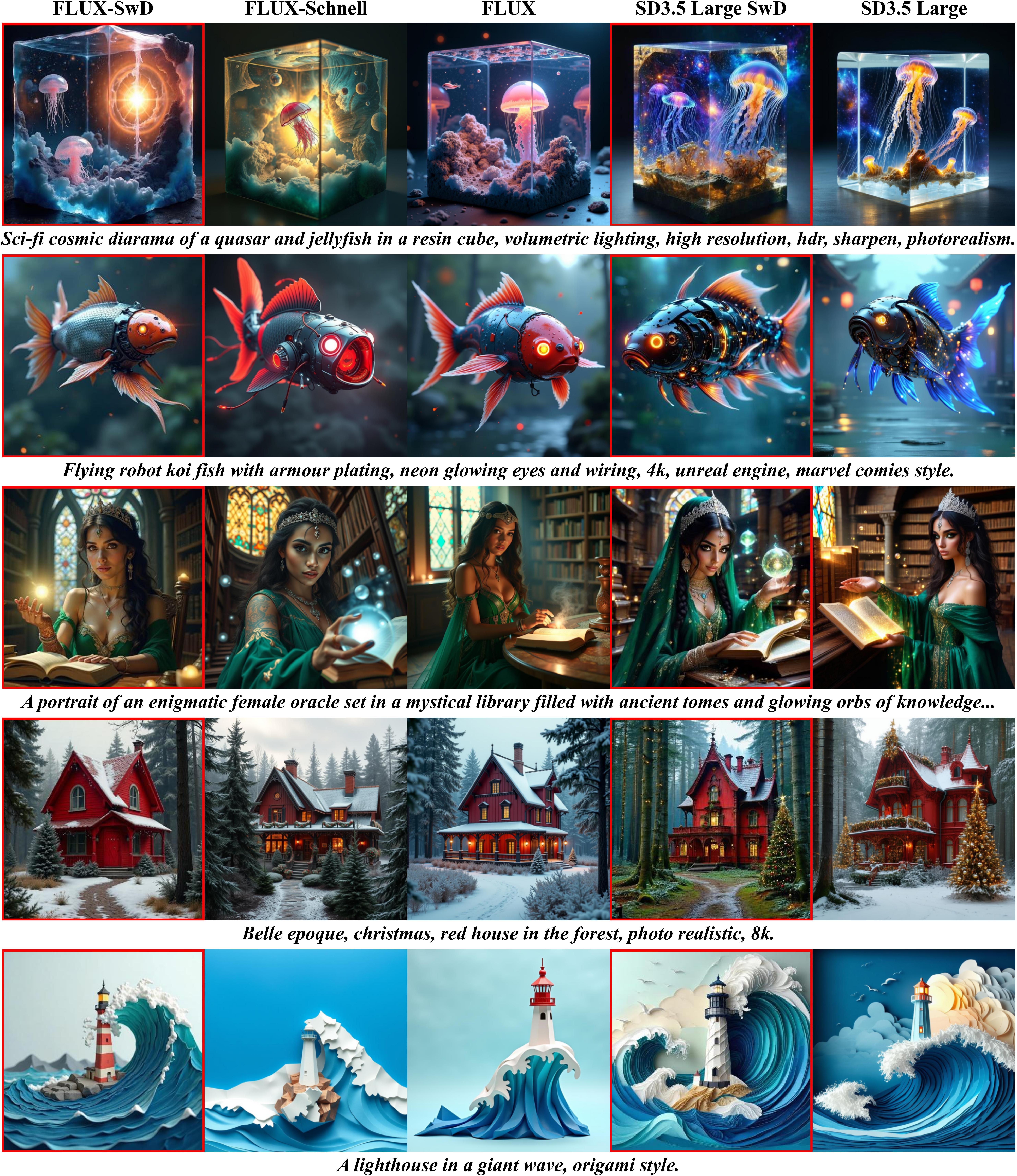}
    \caption{Qualitative results of FLUX-SwD and SD3.5 Large SwD.}
    \label{fig:exps_cherry_app}
\end{figure*}

\begin{figure*}
    \centering
    \includegraphics[width=1.0\linewidth]{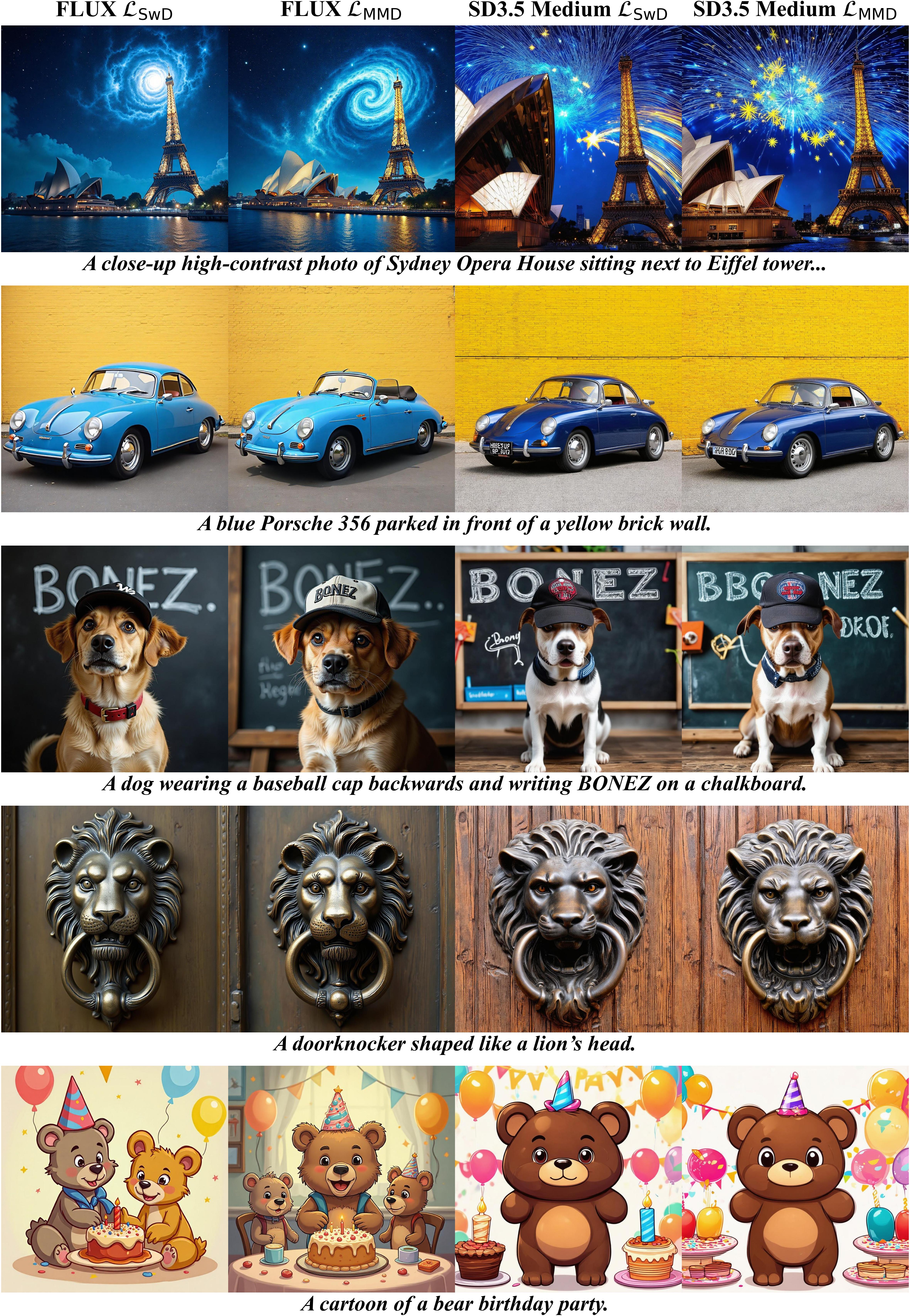}
    \caption{Qualitative comparisons of SwD trained with the full $\mathcal{L}_\text{SwD}$ loss against the ones trained with $\mathcal{L}_\text{MMD}$ alone. $\mathcal{L}_\text{MMD}$ in isolation produces competitive few-step models.}
    \label{fig:app_pdm_vs_swd}
\end{figure*}

\begin{figure*}
    \centering
    \includegraphics[width=1.0\linewidth]{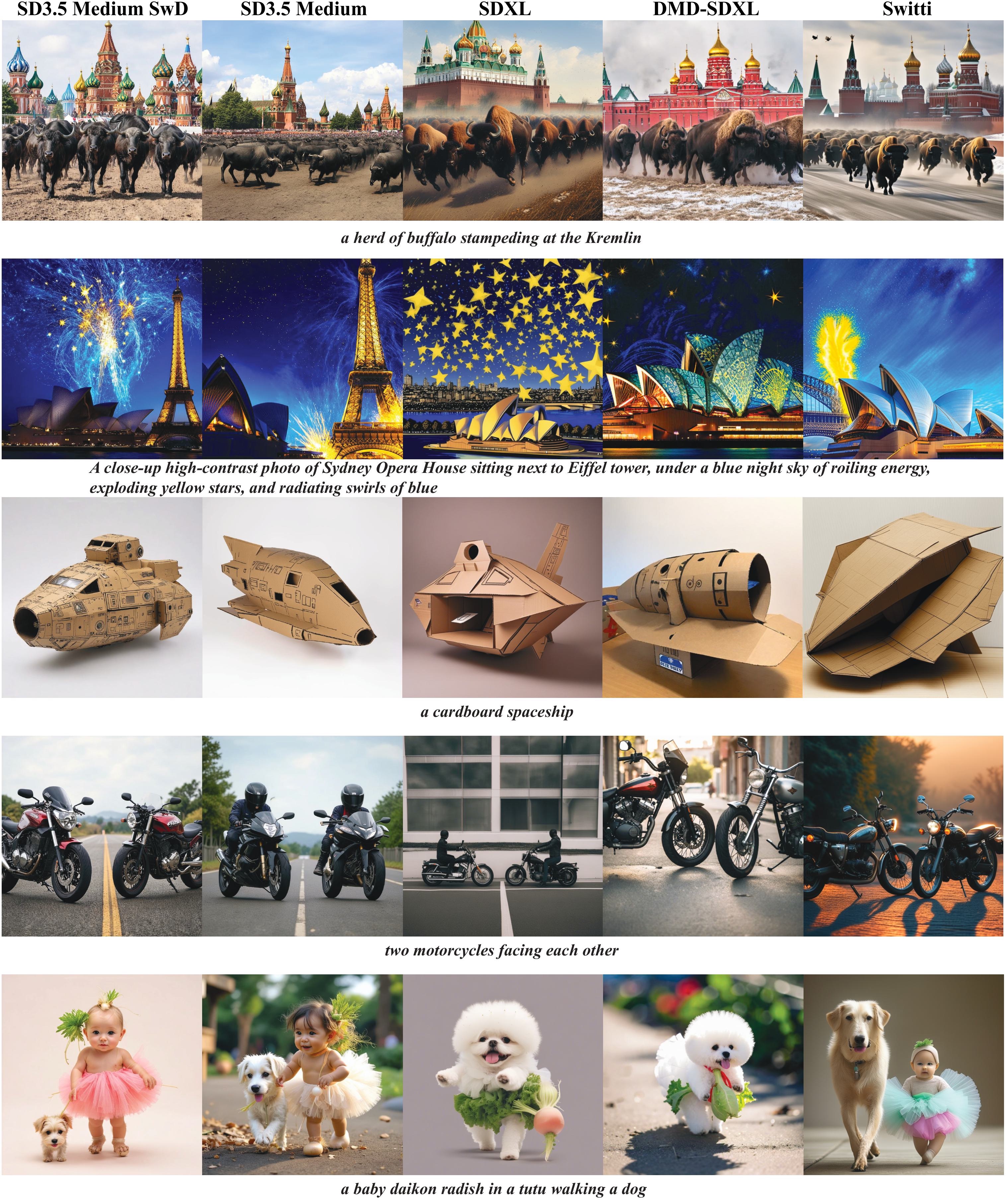}
    \caption{Qualitative comparison of SD3.5 Medium SwD against the models of the similar size.}
    \label{fig:app_cherry}
\end{figure*}

\begin{figure*}
    \centering
    \includegraphics[width=1.0\linewidth]{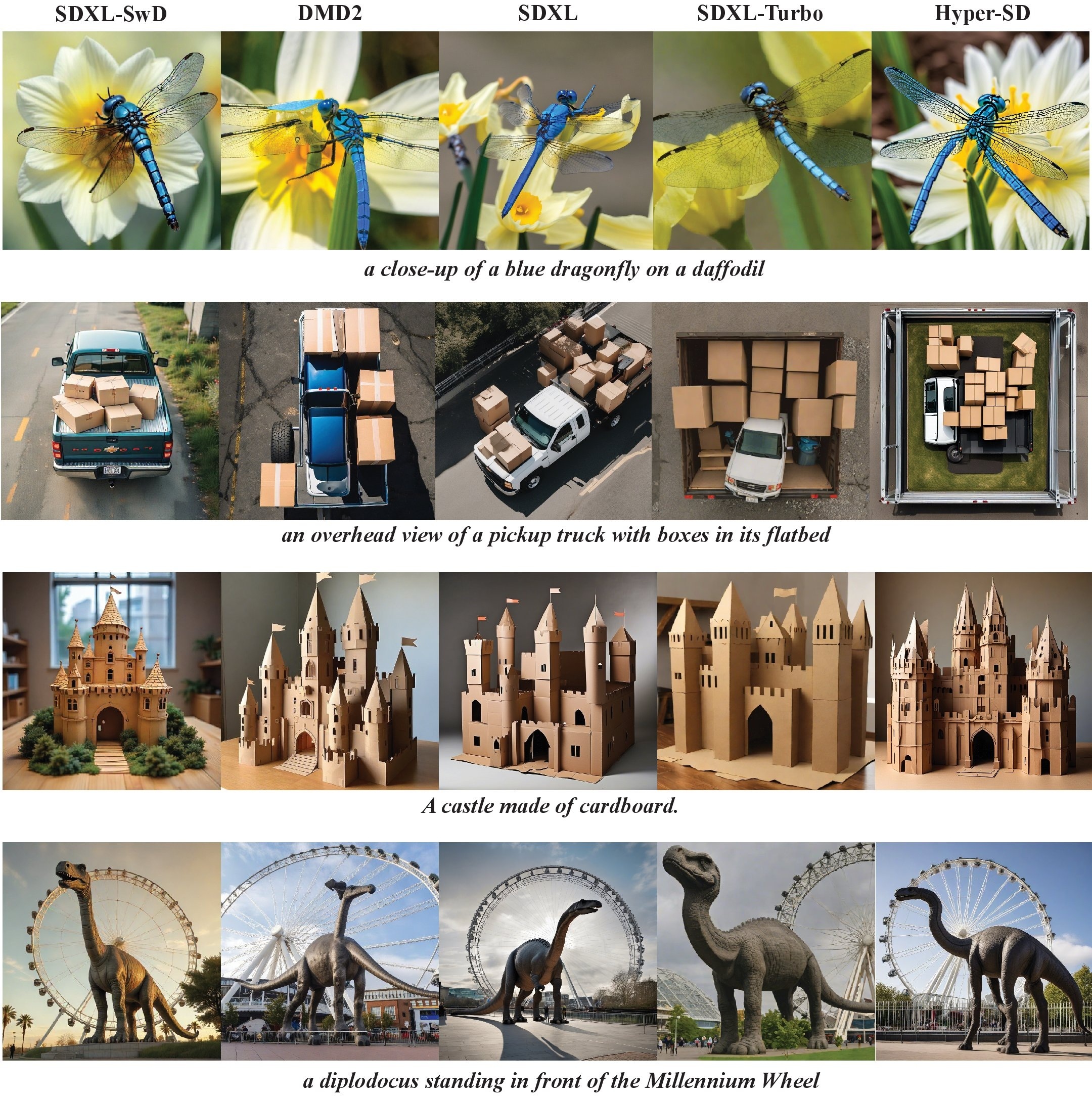}
    \caption{Qualitative comparison of SDXL-SwD against the SDXL-based alternatives.}
    \label{fig:app_sdxl_cherry}
\end{figure*}

%% file: tables/full_vs_scale_coco.tex
\begin{minipage}{0.49\textwidth}
\centering
\resizebox{\linewidth}{!}{%
\begin{tabular}{@{} l *{7}{c} @{}}
\toprule
Setup & Steps & PS $\uparrow$ & HPSv3 $\uparrow$ & IR $\uparrow$ & FID $\downarrow$  \\
\bottomrule
\multicolumn{6}{c}{\rule{0pt}{2.3ex}{SD3.5 Medium}}\\
\midrule
        Scale-wise & $6$ & $\mathbf{22.8}$ & $\mathbf{11.7}$ & ${1.10}$  & $23.1$ \\
        Scale-wise & $4$ & ${22.7}$ & $\mathbf{11.7}$ & $\mathbf{1.12}$ & $23.7$ \\
        Scale-wise & $2$ & ${22.6}$ & ${10.6}$ & ${1.09}$ & $22.3$ \\
        Full-scale & $6$ & ${22.5}$ & ${11.2}$ & ${1.08}$ & $20.4$ \\
        Full-scale & $4$ & ${22.5}$ & ${11.3}$ & ${1.09}$ & $21.2$ \\
        Full-scale & $2$ & ${22.3}$ & ${10.8}$ & ${1.03}$ & $\mathbf{20.3}$ \\
\bottomrule
\multicolumn{6}{c}{\rule{0pt}{2.3ex}{FLUX}}\\
\midrule
        Scale-wise & $4$ & $\mathbf{23.1}$  & $\mathbf{14.6}$ & $\mathbf{1.14}$ & $\mathbf{26.4}$ \\
        Scale-wise & $2$ & ${23.0}$  & ${14.1}$ & ${1.12}$ & $26.5$ \\
        Full-scale & $4$ & $\mathbf{23.1}$ & ${14.0}$ & ${1.13}$ & $28.5$ \\
        Full-scale & $2$ & ${23.0}$ & ${13.8}$ & ${1.13}$ & $26.9$ \\
\bottomrule
\end{tabular}}
\vspace{-2mm}
\caption{
    Quantitative comparison between scale-wise and full-scale setups in terms of automatic metrics on COCO30K.
}
\label{tab:exps_full_vs_scale_coco}
\end{minipage}

%% file: tables/full_vs_scale_mjhq.tex
\begin{minipage}{0.49\textwidth}
\resizebox{\linewidth}{!}{%
\begin{tabular}{@{} l *{7}{c} @{}}
\toprule
Setup & Steps & PS $\uparrow$ & HPSv3 $\uparrow$ & IR $\uparrow$ & FID $\downarrow$  \\
\bottomrule
\multicolumn{6}{c}{\rule{0pt}{2.3ex}{SD3.5 Medium}}\\
\midrule
    Scale-wise & $6$ & $\mathbf{21.8}$ & $\mathbf{10.7}$ & ${1.10}$  & $13.4$ \\
    Scale-wise & $4$ & $\mathbf{21.8}$ & $\mathbf{10.7}$ & $\mathbf{1.13}$ & $13.7$ \\
    Scale-wise & $2$ & ${21.7}$ & ${10.3}$ & ${1.10}$ & $\mathbf{12.8}$ \\
    Full-scale & $6$ & ${21.6}$ & ${10.3}$ & ${1.09}$ & $13.4$ \\
    Full-scale & $4$ & ${21.7}$ & ${10.4}$ & ${1.10}$ & $13.5$ \\
    Full-scale & $2$ & ${21.5}$ & ${10.0}$ & ${1.04}$ & $13.1$ \\
\bottomrule
\multicolumn{6}{c}{\rule{0pt}{2.3ex}{FLUX}}\\
\midrule
    Scale-wise & $4$ & $\mathbf{21.9}$ & $\mathbf{11.6}$ & ${1.06}$ & $14.4$ \\
    Scale-wise & $2$ & $\mathbf{21.9}$ & ${11.5}$ & $\mathbf{1.10}$ & $14.0$ \\
    Full-scale & $4$ & ${21.8}$ & ${11.3}$ & ${1.09}$ & $14.4$ \\
    Full-scale & $2$ & ${21.8}$ & ${11.2}$ & ${1.08}$ & $\mathbf{13.4}$ \\
\bottomrule
\end{tabular}}
\vspace{-2mm}
\caption{
    Quantitative comparison between scale-wise and full-scale setups in terms of automatic metrics on MJHQ30K.
}
\label{tab:exps_full_vs_scale_mjhq}
\end{minipage}

%% file: tables/extended_table1.tex
\begin{wraptable}{l}{0.5\textwidth}
\vspace{2mm}
\centering
\resizebox{0.5\textwidth}{!}{%
\begin{tabular}{@{} l *{5}{c} @{}}
\toprule
Configuration  & $t=400$  & $t=600$ & $t=800$ \\
\midrule
\rowcolor[HTML]{eeeeee} $ \mathbf{x}_0^{128\times128}\xrightarrow[]{\text{noise}}\mathbf{x}_t^{128\times128} $ 
    & $9.2$  & $10.3$ & $13.7$ \\
\midrule
$ \mathbf{x}_0^{32\times32}\xrightarrow[]{\text{upscale}}\mathbf{x}_0^{128\times128} \xrightarrow[]{\text{noise}}\mathbf{x}_t^{128\times128} $ 
    & $90.2$ & $40.8$ & $16.6$ \\
$\mathbf{x}_0^{32\times32}\xrightarrow[]{\text{noise}}\mathbf{x}_t^{32\times32} \xrightarrow[]{\text{upscale}}\mathbf{x}_t^{128\times128} $ 
    & $217$  & $298$  & $361$ \\
\midrule
$ \mathbf{x}_0^{64\times64}\xrightarrow[]{\text{upscale}}\mathbf{x}_0^{128\times128} \rightarrow \mathbf{x}_t^{128\times128}$ 
    & $30.9$ & $18.8$ & $14.7$ \\
$ \mathbf{x}_0^{64\times64}\xrightarrow[]{\text{noise}}\mathbf{x}_t^{64\times64} \xrightarrow[]{\text{upscale}}\mathbf{x}_t^{128\times128}$ 
    & $122$  & $223$  & $327$ \\
\midrule
$ \mathbf{x}_0^{96\times96}\xrightarrow[]{\text{upscale}}\mathbf{x}_0^{128\times128} \xrightarrow[]{\text{noise}}\mathbf{x}_t^{128\times128} $ 
    & $18.8$ & $14.3$ & $14.3$ \\
$\mathbf{x}_0^{96\times96}\xrightarrow[]{\text{noise}}\mathbf{x}_t^{96\times96} \xrightarrow[]{\text{upscale}}\mathbf{x}_t^{128\times128} $ 
    & $32.8$ & $58.9$ & $120$ \\
\bottomrule
\end{tabular}}
\caption{
    Extended results of the noisy latent upscaling strategies. 
    We evaluate FID-5K on COCO2014 by generating images with SD3.5-M using $\mathbf{x}_t$ obtained with different upsampling strategies discussed in~\Cref{sect:method_scalewise}.
}
\label{tab:extended_table1}
\end{wraptable}